\documentclass[letterpaper]{article}

\usepackage{aaai2027}
\usepackage{times}
\usepackage{helvet}
\usepackage{courier}
\usepackage{graphicx}
\usepackage{enumitem}
\usepackage{amsmath}
\usepackage{amssymb}
\usepackage{natbib}
\usepackage{multirow}
\usepackage{booktabs}
\usepackage{algorithm}
\usepackage{algorithmic}
\usepackage[hyphens]{url}
\usepackage{fontawesome5}

\usepackage{eso-pic}

\usepackage[table]{xcolor}

\definecolor{bestbg}{RGB}{235,224,252}
\definecolor{secondbg}{RGB}{255,238,246}
\definecolor{backbonebg}{RGB}{245,245,245}

\newcommand{\best}[1]{\cellcolor{bestbg}\textbf{#1}}
\newcommand{\second}[1]{\cellcolor{secondbg}\underline{#1}}

\usepackage{listings}
\usepackage[most]{tcolorbox}
\usepackage{needspace}
\usepackage{cuted}

\DeclareRobustCommand{\bestcap}[1]{%
  \begingroup
  \setlength{\fboxsep}{1.2pt}%
  \colorbox{bestbg}{\strut\textbf{#1}}%
  \endgroup
}

\DeclareRobustCommand{\secondcap}[1]{%
  \begingroup
  \setlength{\fboxsep}{1.2pt}%
  \colorbox{secondbg}{\strut\underline{#1}}%
  \endgroup
}

\definecolor{darkpink}{RGB}{210,90,145}

\usepackage[
    colorlinks=true,
    citecolor=darkpink,
    linkcolor=darkpink,
    urlcolor=darkpink
]{hyperref}

\definecolor{emailpurple}{RGB}{105,75,145}

\nocopyright

\begin{document}

% % 仅在第一页添加左上角 Logo 和横线
% \AddToShipoutPictureFG*{%
%   \AtPageUpperLeft{%

%     % Logo
%     % 0.55in：距离页面左侧
%     % 0.42in：距离页面顶部
%     \put(\LenToUnit{0.55in},\LenToUnit{-0.42in}){%
%       \raisebox{-\height}{%
%         \includegraphics[
%           height=0.3in,
%           keepaspectratio
%         ]{Figures/wenxin_1.png}%
%       }%
%     }%

%     % 右上角日期
%     \put(\LenToUnit{0.48in},\LenToUnit{-0.74in}){%
%       \makebox[\dimexpr\paperwidth-0.96in\relax][r]{%
%         \large\color{gray!75!black}August 3, 2026%
%       }%
%     }%

%     % Logo 下方横线
%     % 横线距离页面顶部约 1.18in
%     \put(\LenToUnit{0.48in},\LenToUnit{-0.8in}){%
%       \rule{\dimexpr\paperwidth-0.96in\relax}{1.0pt}%
%     }%
%   }%
% }

% \title{Supplementary Material:
% Group-Reflective Self-Distillation for Agentic Reinforcement Learning}

% \author{
% Anonymous Authors
% }

\title{Group-Reflective Self-Distillation for Agentic Reinforcement Learning}

\author{
    Binbin Zheng\textsuperscript{\rm 1,\rm 2}%
    \thanks{Equal contribution.}%
    \thanks{This work was done during an internship at Baidu.},
    Zijun Xie\textsuperscript{\rm 2,\rm 3}%
    \footnotemark[1]\footnotemark[2],
    Guanqun Zhao\textsuperscript{\rm 2,\rm 4}%
    \footnotemark[1]\footnotemark[2],
    Enlei Gong\textsuperscript{\rm 2}\corresponding\\
    Xing Ma\textsuperscript{\rm 5},
    Xiaoliang Fu\textsuperscript{\rm 6},
    Zeyu Chen\textsuperscript{\rm 2}\corresponding
}

\affiliations{
    \textsuperscript{\rm 1}University of Science and Technology of China\quad
    \textsuperscript{\rm 2}Baidu Inc.\quad
    \textsuperscript{\rm 3}Peking University\\
    \textsuperscript{\rm 4}Beijing University of Posts and Telecommunications\quad
    \textsuperscript{\rm 5}Tianjin University\quad
    \textsuperscript{\rm 6}Fudan University\\[0.5em]
    {\small\ttfamily\color{emailpurple}
      % binbinzheng@mail.ustc.edu.cn
      zhengbinbin01@baidu.com
    }
    \href{https://github.com/BinbZheng1/GRSD}{%
      \faGithub\ \texttt{https://github.com/BinbZheng1/GRSD}%
    }
}

\maketitle

\begin{abstract}
Reinforcement learning with verifiable rewards (RLVR) is effective for training large language model agents. However, terminal rewards provide only coarse trajectory-level supervision, leaving successful behaviors, recurring mistakes, and incidental choices entangled in the same outcome signal. Existing agentic self-distillation methods enrich sparse supervision with natural-language skills, but skills retrieved externally or extracted from a single trajectory by stronger models may mismatch current experience, exceed the policy's capability, or remain path-specific. We propose Group-Reflective Self-Distillation (GRSD), which derives capability-aligned and outcome-discriminative guidance from the policy's own verified rollouts. For each prompt, the policy reflects on each verified trajectory in an on-policy group, and a stop-gradient snapshot contrasts the resulting reflections from successful and failed rollouts to construct group-level privileged guidance. Conditioned on this guidance, a self-teacher refines turn-level credit assignment by modulating outcome-based advantages while preserving the verifier-determined learning direction. Experiments across multiple agentic environments and model scales demonstrate that GRSD consistently outperforms competitive baselines and generalizes more effectively to unseen tasks.
\end{abstract}

\section{Introduction}
Large language model (LLM) agents solve complex tasks through multi-turn interaction with tools and environments~\citep{yao2022react, schick2023toolformer}. Unlike single-turn reasoning models, agents repeatedly take actions, receive observations, and adapt their decisions to evolving states. Reinforcement learning with verifiable rewards (RLVR) therefore provides a natural training paradigm for agentic tasks~\citep{dong2025agentic, feng2026group, jin2025search}. In particular, outcome-based methods such as GRPO~\citep{shao2024deepseekmath} optimize agent policies directly from task success or failure without requiring a learned critic. However, terminal rewards provide only coarse trajectory-level supervision, leaving successful behaviors, recurring mistakes, and incidental trajectory choices entangled in the same outcome signal~\citep{tan2026hindsight, xie2026echo}.  Consequently, outcome supervision alone provides limited guidance on which behaviors should be reinforced or corrected during policy learning.

On-policy self-distillation~\citep{zhao2026self, yang2026self, hubotter2026reinforcement, pan2026rlcsd} offers a promising way to enrich such sparse supervision by treating the current policy conditioned on privileged information as a self-teacher. Recent agentic extensions, including Skill-SD~\citep{wang2026skill} and SDAR~\citep{lu2026self}, further incorporate natural-language skills into multi-turn policy optimization. These methods typically retrieve relevant skills from an externally constructed skill repository and use them as privileged context. Although retrieval enables the reuse of general behavioral knowledge, the selected skills may not accurately match the current prompt, interaction trajectory, or environment state. Consequently, retrieval-based guidance can remain only loosely aligned with the agent's current on-policy experience.

\begin{figure}[t]
\centering
\includegraphics[width=0.98\linewidth]{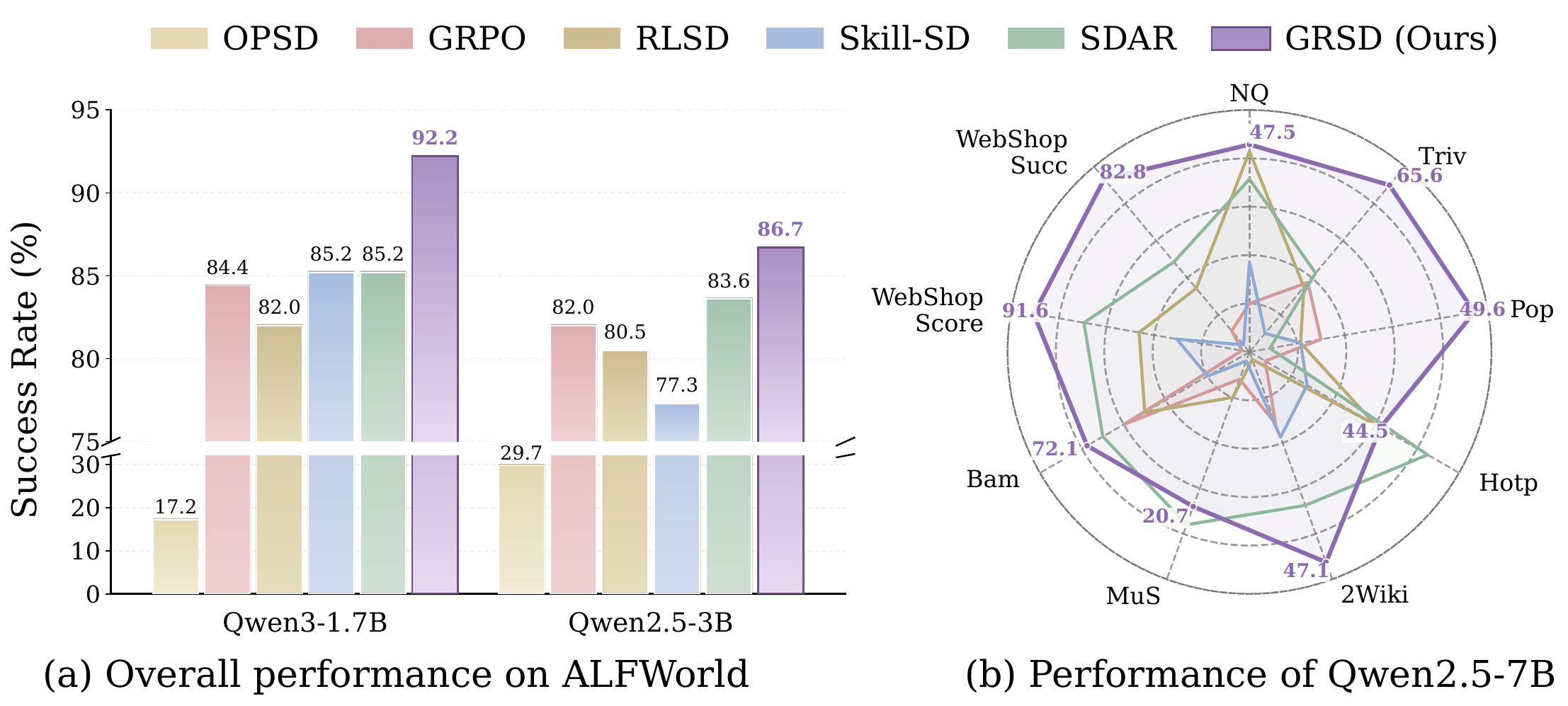}
\caption{Performance overview of GRSD across multiple agentic environments and backbones.}
\label{fig:1}
\end{figure}

A recent approach alleviates this mismatch by extracting hindsight skills directly from completed on-policy trajectories~\citep{yang2026opid}. 
However, these trajectories are typically analyzed or summarized by a stronger external model, yielding guidance that may exceed the current policy's capability boundary and fail to reflect its own ability to abstract reusable skills from experience.
Moreover, the resulting skills are typically extracted from each trajectory independently, without contrasting multiple successful and failed rollouts under the same prompt. 
Since a successful trajectory may contain unnecessary detours and a failed trajectory may still include useful behaviors, single-trajectory skills may capture path-specific details rather than the behavioral patterns that truly distinguish success from failure.
These limitations raise two key challenges. First, \textbf{how can the policy derive capability-aligned guidance from its own on-policy experience?} Second, \textbf{how can it contrast rollouts to identify success-critical behaviors and recurring failure modes?}

To address these challenges, we propose \textsc{GRSD}, \textbf{G}roup-\textbf{R}eflective \textbf{S}elf-\textbf{D}istillation for agentic reinforcement learning. For each prompt, \textsc{GRSD} samples a group of on-policy rollouts and performs self-reflection on each verified trajectory. Specifically, the policy identifies outcome-critical decisions, explains how they contribute to success or failure, and summarizes concrete behaviors to reinforce or avoid. A stop-gradient snapshot of the same policy then contrasts reflections from successful and failed rollouts and synthesizes them into compact group-reflective guidance. This policy-native construction better aligns the resulting guidance with the capability of the policy being optimized, while the success--failure contrast encourages it to capture outcome-discriminative behavioral patterns. During optimization, the group-reflective guidance serves as privileged context for a self-teacher that assigns turn-specific credit by modulating outcome-based advantages without changing their verifier-determined signs. By enabling the policy to derive reusable guidance from its own verified experience and internalize it during optimization, \textsc{GRSD} establishes an experience-driven loop for agentic self-evolution.

Our main contributions are as follows.
\begin{itemize}[leftmargin=1em]
\item We propose \textsc{GRSD}, which learns policy-native reflections from verified trajectories and contrasts successful and failed experiences to derive group-level guidance without external models for skill generation.

\item We develop a self-distillation mechanism that leverages group-reflective guidance to refine trajectory-level advantages into turn-specific credit while preserving the verifier-determined learning direction.

\item Extensive evaluations across diverse agentic environments and model scales demonstrate that \textsc{GRSD} achieves leading aggregate performance and generalizes effectively to unseen tasks.
\end{itemize}

\section{Related Work}

\subsection{Agentic Reinforcement Learning for LLMs}
LLMs are increasingly deployed as interactive agents that reason over long horizons, invoke tools, and adapt to environmental feedback~\citep{yao2022react, schick2023toolformer}. 
Reinforcement learning with verifiable rewards (RLVR), particularly critic-free methods such as GRPO~\citep{shao2024deepseekmath}, directly optimizes agent policies from verified task outcomes and has become a widely used post-training paradigm~\citep{chen2025minimax, fu2026log, zhao2026deconstructingoffpolicyratiosentropyscaled, fu2026maspo, dong2025agentic, feng2026group, jin2025search}. 
However, sparse and delayed terminal rewards indicate only whether a trajectory succeeds or fails, without identifying the intermediate decisions responsible for the outcome~\citep{tan2026hindsight, xie2026echo}. 
As a result, effective behaviors, recurring mistakes, and incidental choices remain entangled in the same trajectory-level signal. 
\textsc{GRSD} retains outcome-based RL as the primary objective and uses the policy's own verified rollouts to refine trajectory-level advantages into turn-specific credit without altering the verifier-determined learning direction.

\subsection{On-Policy Distillation and Self-Distillation}
On-policy distillation (OPD) densifies sparse rewards by supervising the student's own samples with teacher token-level signals~\citep{agarwal2024policy, gu2024minillm, li2026rethinking, zheng2026scope, xu2026tip, zhang2026reinforcement, zhong2026sod}. 
However, it typically relies on a separate stronger teacher, requiring access to teacher logits and a shared vocabulary~\citep{ye2025black, sun2026simct, fu2026revisiting}. 
On-policy self-distillation~\citep{zhao2026self, yang2026self} removes this requirement by using a single model as both the privileged-context teacher and the plain-context student. 
For multi-turn agents, recent work further distills natural-language \emph{skills} into the policy~\citep{wang2026skill, lu2026self, yang2026opid}. 
These skills describe subgoals, reusable action patterns, or behavioral rules, and mainly differ in their source: external skill memories~\citep{wang2026skill, lu2026self} or hindsight extraction from completed trajectories using an external analyzer~\citep{yang2026opid}. 
The former may mismatch the current state or interaction history, while the latter may introduce guidance beyond the capability of the policy being optimized when the analyzer is much stronger. 
Moreover, trajectory-derived skills are typically extracted from individual rollouts without contrasting successful and failed experiences under the same prompt, making them susceptible to path-specific details. 
In contrast, \textsc{GRSD} derives \emph{policy-native}, \emph{group-reflective} guidance by contrasting the policy's own reflections on successful and failed rollouts for the same prompt. 
This construction yields capability-aligned and outcome-discriminative training guidance without relying on external skill libraries or external models for skill generation, and requires no privileged context at inference.

\section{Methodology}
\label{sec:methodology}

We first formulate multi-turn agentic RL and GRPO in Section~\ref{sec:preliminaries}. Section~\ref{sec:skill_prior} constructs group-reflective guidance from policy-native trajectory reflections. Section~\ref{sec:turn_level_distillation} leverages this guidance to refine turn-level credit, and Section~\ref{sec:objective} presents the joint task and reflection objective.

\begin{figure*}[t]
    \centering
    \includegraphics[width=\linewidth]{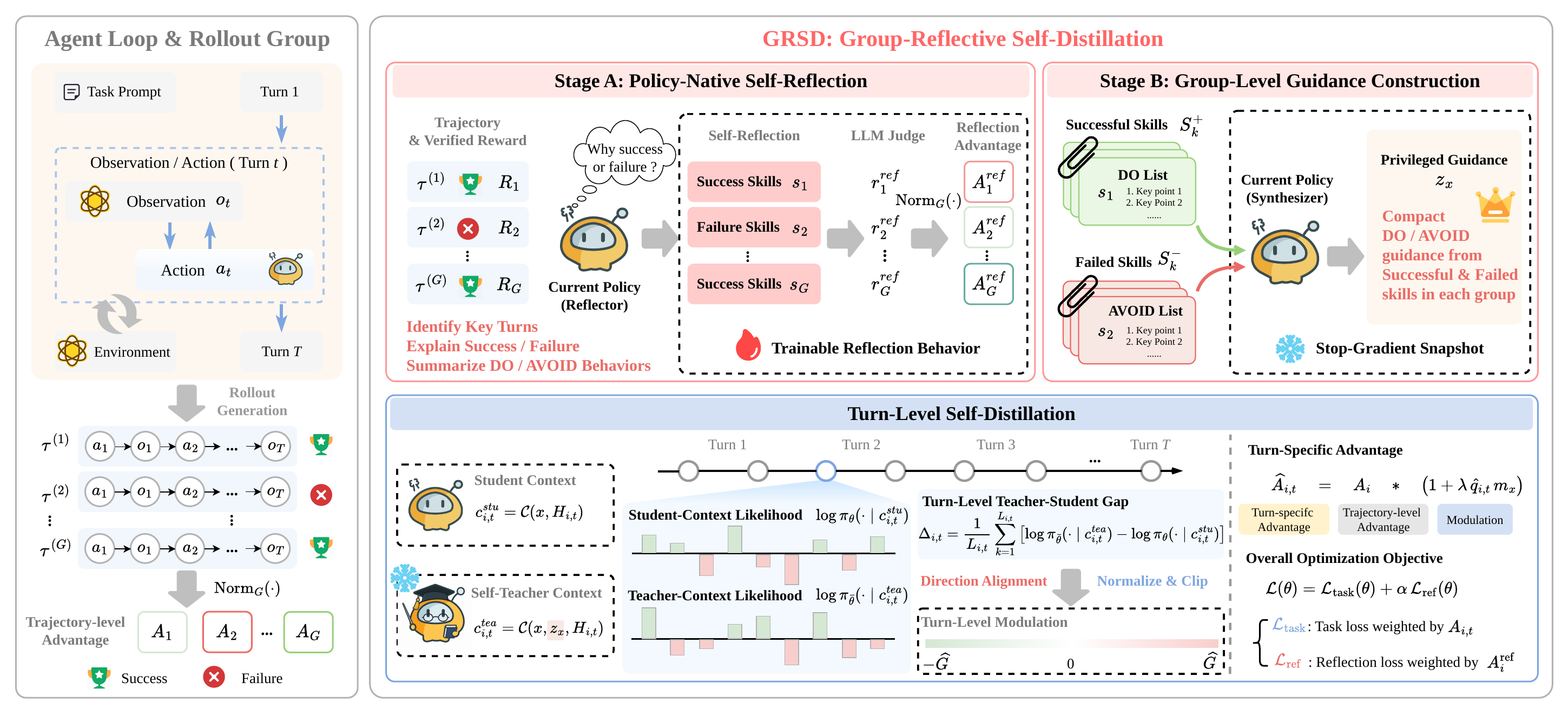}
    \caption{Overview of GRSD. Stage~A learns policy-native reflection skills from verified trajectories. Stage~B contrasts skills from successful and failed rollouts to construct group-level DO/AVOID guidance. Conditioned on this privileged guidance, a self-teacher converts the trajectory-level advantage into turn-level modulated advantages without reversing its direction.}
    \label{fig:grsd}
\end{figure*}

\subsection{Preliminaries}
\label{sec:preliminaries}

We consider a tool-use agent that interacts with an environment over multiple turns. Let $x$ denote the initial prompt. At turn $t$, the agent conditions on $c_t=\mathcal{C}(x,H_t)$ and samples an action $a_t=(y_{t,1},\ldots,y_{t,L_t})\sim\pi_\theta(\cdot\mid c_t)$, after which the environment returns an observation $o_t$. Here, $H_t=((a_1,o_1),\ldots,(a_{t-1},o_{t-1}))$, and the complete trajectory is $\tau=((a_1,o_1),\ldots,(a_T,o_T))$. The context function $\mathcal{C}$ is shared by all compared methods.

For each prompt $x$, GRPO samples a group of $G$ rollouts
$\{\tau_i\}_{i=1}^{G}$ from the old policy $\pi_{\theta_{\mathrm{old}}}$.
After each trajectory is completed, a verifier assigns a task reward
$R_i=R(\tau_i)$ and an outcome label $b_i\in\{0,1\}$ indicating failure or success. For any group-level scalar signal $v_i$, we define
\begin{equation}
\label{eq:group_norm}
    \operatorname{Norm}_G(v_i)
    =
    \frac{v_i-\bar{v}}
    {
    \sqrt{
    \frac{1}{G}\sum_{j=1}^{G}(v_j-\bar{v})^2+\epsilon_{\mathrm{adv}}
    }
    },\;
    \bar{v}=\frac{1}{G}\sum_{j=1}^{G}v_j .
\end{equation}
The trajectory-level GRPO advantage is then computed from the outcome rewards as
$A_i=\operatorname{Norm}_G(R_i)$.

The standard GRPO objective applies this trajectory-level advantage to every
action token in the same rollout:
\begin{equation}
\label{eq:grpo}
\begin{aligned}
    \mathcal{L}_{\mathrm{GRPO}}(\theta)
    &=
    -\mathbb{E}
    \Bigg[
    \frac{1}{G}
    \sum_{i=1}^{G}
    \frac{1}{N_i}
    \sum_{t=1}^{T_i}
    \sum_{k=1}^{L_{i,t}}
    \min
    \Big(
    \rho_{i,t,k}A_i,\\
    &\qquad\qquad
    \mathrm{clip}(\rho_{i,t,k},1-\varepsilon,1+\varepsilon)A_i
    \Big)
    \Bigg],
\end{aligned}
\end{equation}
\noindent where $L_{i,t}$ is the number of action tokens at turn $t$,
$N_i=\sum_{t=1}^{T_i}L_{i,t}$ is the total number of action tokens in $\tau_i$, and
\[
\rho_{i,t,k}
=
\frac{
\pi_\theta(y_{i,t,k}\mid c_{i,t},y_{i,t,<k})
}{
\pi_{\theta_{\mathrm{old}}}(y_{i,t,k}\mid c_{i,t},y_{i,t,<k})
}.
\]
For clarity, the shared KL regularization toward the fixed reference policy is
omitted from all displayed objectives. Since the same $A_i$ is assigned to all
action tokens in $\tau_i$, vanilla GRPO does not distinguish how individual
tool calls, observation-dependent decisions, or intermediate reasoning steps
contribute to the final outcome.

\subsection{Group-Reflective Guidance Construction}
\label{sec:skill_prior}

\textsc{GRSD} constructs privileged guidance in two stages. Stage~A addresses
capability alignment by generating a policy-native reflection from each
verified trajectory, whose content serves as a trajectory-level skill.
Stage~B addresses path specificity by contrasting skills from successful and
failed rollouts of the same prompt and synthesizing compact,
outcome-discriminative guidance.

\paragraph{Stage A: Policy-Native Self-Reflection.}
After trajectory $\tau_i$ is completed and its outcome $b_i$ is verified, the
same rollout policy follows the reflection instruction $p_{\mathrm{ref}}$ to
extract a trajectory-level skill, formally expressed as:
\begin{equation}
\label{eq:skill_reflection}
    s_i
    \sim
    \pi_{\theta_{\mathrm{old}}}
    \left(
    \cdot
    \mid
    x,\tau_i,b_i,p_{\mathrm{ref}}
    \right).
\end{equation}
The instruction elicits a structured reflection that identifies
outcome-critical turns, relates them to the final verified outcome, and
distills concrete behaviors to reinforce or avoid. Since $s_i$ is generated
by a snapshot of the policy being optimized rather than by a stronger external
analyzer, it remains policy-native and aligned with the policy's current
skill-induction capability. The reflection objective in
\S\ref{sec:objective} further improves the quality of these
reflection-derived skills.

To supervise reflection quality, a fixed external judge $\mathcal{J}$ assigns
each reflection a scalar score:
\begin{equation}
\label{eq:reflection_reward}
    r^{\mathrm{ref}}_i
    =
    \mathcal{J}(s_i,\tau_i,b_i)
    \in \{0,1,2,3\}.
\end{equation}
The judge evaluates whether the identified turns are consistent with the
verified outcome and whether the resulting reflection is specific,
well-grounded, and actionable. It provides only this scalar score and never
generates or revises the reflection content. We compute the reflection
advantage using the same normalization operator as
Eq.~(\ref{eq:group_norm}):
$A^{\mathrm{ref}}_i=\operatorname{Norm}_G(r^{\mathrm{ref}}_i)$.
This provides the reflection tokens with their own learning signal, distinct
from the task-execution reward.

\paragraph{Stage B: Group-Level Guidance Construction.}
For each prompt $x$, we partition the reflection-derived skills according to
their verified outcome rewards:
\begin{equation}
\label{eq:skill_sets}
\begin{aligned}
    \mathcal{S}^{+}_x &= \{s_i \mid b_i=1\},&
    \mathcal{S}^{-}_x &= \{s_i \mid b_i=0\}.
\end{aligned}
\end{equation}
Let $\bar\theta=\operatorname{sg}(\theta_{\mathrm{old}})$ denote the
stop-gradient rollout-policy snapshot. It contrasts the two skill sets and
synthesizes prompt-specific guidance:
\begin{equation}
\label{eq:skill_prior}
    z_x
    \sim
    \pi_{\bar\theta}
    \left(
    \cdot
    \mid
    x,\mathcal{S}^{+}_x,\mathcal{S}^{-}_x,p_{\mathrm{guide}}
    \right).
\end{equation}
The instruction $p_{\mathrm{guide}}$ asks the policy to retain recurring
behaviors associated with success, identify mistakes associated with failure,
and organize them into a concise DO/AVOID guide. By contrasting skills across
the two outcome groups, $z_x$ emphasizes behavioral differences that are more
likely to be outcome-critical while suppressing idiosyncratic details from
individual trajectories. Stage~A is learned through reflection rewards,
whereas Stage~B is a detached construction: no gradient flows through the
sampled guidance $z_x$, which is treated as fixed privileged context for
subsequent turn-level credit refinement. The two stages are treated
differently because training Stage~A develops a reusable ability to extract
actionable skills from experience, which may facilitate the emergence of
stronger task-solving capabilities, whereas detaching Stage~B prevents a
prompt-specific aggregation objective from competing with task execution.

We enable guidance-based credit refinement only for prompt groups that contain
both successful and failed rollouts:
\begin{equation}
    m_x
    =
    \mathbf{1}
    \left[
    |\mathcal{S}^{+}_x|>0
    \wedge
    |\mathcal{S}^{-}_x|>0
    \right].
\end{equation}
When either set is empty, we set $m_x=0$, and
Eq.~(\ref{eq:grsd_advantage}) reduces to $\widehat A_{i,t}=A_i$; the task
objective therefore falls back to standard GRPO. The guidance is used only to
construct training signals and is never provided at inference time.

\subsection{Turn-Level Self-Distillation}
\label{sec:turn_level_distillation}

Following prior work on turn-level credit assignment in agentic
RL~\citep{zhong2026sod}, we use the group-reflective guidance as privileged
information to refine the trajectory-level GRPO advantage into turn-specific
credit. For each sampled turn, we construct a plain student context and a
guidance-augmented teacher context:
\begin{equation}
    c^{\mathrm{stu}}_{i,t}=\mathcal{C}(x,H_{i,t}),
    \qquad
    c^{\mathrm{tea}}_{i,t}=\mathcal{C}(x,z_x,H_{i,t}).
\end{equation}
The same stop-gradient policy $\pi_{\bar\theta}$ evaluates the \emph{same}
sampled action tokens under both contexts. Their mean log-likelihood difference
quantifies how the group-reflective guidance changes
the policy's support for the sampled turn:
\begin{equation}
\label{eq:skill_gap}
\begin{aligned}
    \Delta_{i,t}
    &=
    \frac{1}{L_{i,t}}
    \sum_{k=1}^{L_{i,t}}
    \Bigl[
    \log \pi_{\bar\theta}
    \left(y_{i,t,k}\mid y_{i,t,<k},c^{\mathrm{tea}}_{i,t}\right)
    \\
    &\qquad\qquad-
    \log \pi_{\bar\theta}
    \left(y_{i,t,k}\mid y_{i,t,<k},c^{\mathrm{stu}}_{i,t}\right)
    \Bigr].
\end{aligned}
\end{equation}
We align this difference with the outcome-derived update direction by defining
$q_{i,t}=\mathrm{sign}(A_i)\Delta_{i,t}$. For a successful rollout, a turn is
upweighted when guidance increases its likelihood; for a failed rollout, it is
penalized more strongly when guidance decreases its likelihood. A conflicting
teacher signal instead reduces the update magnitude.

We map $q_{i,t}$ to a zero-centered relative likelihood change, normalize it by
the mean absolute magnitude over all valid turns in the same prompt group, and
clip the result to obtain a bounded modulation score:
\begin{equation}
\label{eq:normalized_gap}
    \widehat{q}_{i,t}
    =
    \mathrm{clip}\!\left(
    \frac{\exp(q_{i,t})-1}
    {\frac{1}{|\mathcal{T}_x|}\sum_{(j,t')\in\mathcal{T}_x}
    |\exp(q_{j,t'})-1|+\epsilon_{\mathrm{gap}}},
    -\widehat{G},\widehat{G}
    \right)
\end{equation}
where $\mathcal{T}_x$ contains all valid action turns from the rollouts of
prompt $x$, and $\epsilon_{\mathrm{gap}}>0$ is a numerical stabilizer.

Finally, we leverage this score to refine the trajectory-level advantage into a
turn-specific advantage:
\begin{equation}
\label{eq:grsd_advantage}
    \widehat{A}_{i,t}
    =
    A_i
    \left(
    1+\lambda\,\widehat{q}_{i,t}\,m_x
    \right),
\end{equation}
where $\lambda$ controls the self-distillation strength and $m_x$ is the
mixed-outcome mask defined above. We constrain
$\lambda\widehat{G}\leq 1$ so that the modulation cannot reverse the sign of
$A_i$. As a result, the verifier determines the update direction, while the
privileged branch adjusts the turn-specific advantage according to
each turn's estimated contribution to the final outcome.

\begin{table*}[t]
\centering
\resizebox{\textwidth}{!}{%
\begin{tabular}{l|ccccccc|cccccccc|cc}
\toprule
\multirow{2}{*}{Method}
& \multicolumn{7}{c|}{ALFWorld}
& \multicolumn{8}{c|}{Search-based QA}
& \multicolumn{2}{c}{WebShop}\\
\cmidrule(lr){2-8}\cmidrule(lr){9-16}\cmidrule(lr){17-18}
& Pick & Look & Clean & Heat & Cool & Pick2 & Avg
& NQ & Triv & Pop & Hotp & 2Wiki & MuS & Bam & Avg
& Score & Succ\\
\midrule
\rowcolor{backbonebg}
\multicolumn{18}{c}{\textit{Backbone: Qwen3-1.7B}}\\
\midrule
Vanilla
& 21.9 & 33.3 & 0.0 & 9.1 & 4.5 & 5.9 & 11.7
& 26.9 & 43.2 & 33.6 & 23.2 & 20.1 & 6.4 & 8.9 & 29.2
& 43.3 & 4.7\\

Skill-Prompt
& 28.2 & 50.0 & 8.7 & 0.0 & 4.8 & 0.0 & 14.8
& 27.9 & 46.9 & 36.7 & 23.9 & 20.0 & 6.7 & 9.3 & 30.4
& 30.7 & 6.3\\

OPSD
& 28.2 & 46.7 & 4.8 & 0.0 & 0.0 & 12.5 & 17.2
& 0.7 & 0.6 & 0.4 & 0.4 & 0.9 & 0.0 & 0.0 & 0.5
& 45.5 & 7.8\\

GRPO
& 86.5 & \second{91.7} & 88.9 & \second{90.0} & \second{85.0}
& 68.2 & 84.4
& \best{43.4} & \second{59.9} & 46.3 & \best{38.8}
& 33.6 & \best{13.2} & 64.9 & \second{43.7}
& \second{82.2} & 64.4\\

RLSD
& 81.1 & 84.6 & \best{100.0} & 84.2 & \best{87.0}
& 55.6 & 82.0
& \second{42.7} & 59.3 & 46.1 & \second{38.7}
& \second{35.0} & \second{13.1} & 65.3 & \second{43.7}
& 77.0 & 57.0\\

Skill-SD
& 88.9 & 77.8 & 88.9 & 83.3
& 73.7 & \best{89.5} & \second{85.2}
& 42.1 & 57.5 & \second{46.5} & 36.7
& 33.3 & 12.1 & 65.3 & 42.7
& \best{84.5} & \second{65.6}\\

SDAR
& \best{93.5} & 78.6 & \second{95.7} & 81.0 & 80.0
& 73.7 & \second{85.2}
& 41.9 & 57.6 & \best{46.9} & 36.9
& 34.0 & \second{13.1} & \second{65.7} & 43.1
& 80.0 & 63.3\\

\textbf{GRSD (Ours)}
& \second{92.5} & \best{100.0} & \best{100.0}
& \best{91.3} & 81.3 & \second{88.2} & \best{92.2}
& 41.8 & \best{60.0} & 45.8 & 38.1
& \best{36.4} & 12.2 & \best{66.1} & \best{43.8}
& \best{84.5} & \best{66.4}\\

\midrule
\rowcolor{backbonebg}
\multicolumn{18}{c}{\textit{Backbone: Qwen2.5-3B-Instruct}}\\
\midrule
Vanilla
& 60.0 & 50.0 & 6.3 & 0.0 & 0.0 & 9.1 & 19.5
& 25.2 & 47.4 & 31.8 & 25.8 & 25.8 & 7.6 & 62.1 & 31.6
& 4.9 & 1.6\\

Skill-Prompt
& \second{74.2} & 31.3 & 4.8 & 11.8 & 0.0 & 13.6 & 26.6
& 27.7 & 39.5 & 30.2 & 20.1 & 17.6 & 4.6 & 9.3 & 25.7
& 1.2 & 0.0\\

OPSD
& 56.7 & 42.9 & 17.4 & 10.0 & 17.2 & 27.6 & 29.7
& 0.1 & 0.3 & 0.8 & 0.1 & 0.6 & 0.0 & 0.0 & 0.3
& 11.6 & 2.3\\

GRPO
& \best{100.0} & \best{81.8} & \best{95.7} & 58.8
& 70.6 & 65.2 & 82.0
& 38.5 & 57.9 & \best{43.9} & 36.4
& 35.3 & 12.0 & \second{64.1} & 42.1
& 83.2 & 65.6\\

RLSD
& \best{100.0} & 76.9 & 90.5 & 69.2
& 61.5 & \second{75.0} & 80.5
& 41.2 & 57.8 & 42.2 & 36.4
& 34.4 & 13.6 & \best{64.9} & 41.8
& 84.2 & \second{71.9}\\

Skill-SD
& \best{100.0} & 66.7 & 82.6 & \second{85.7}
& 63.0 & 57.1 & 77.3
& 41.7 & 58.7 & \second{43.5} & 37.6
& 36.9 & 13.4 & \best{64.9} & 43.2
& 77.3 & 65.6\\

SDAR
& \best{100.0} & 66.7 & \second{93.3} & 76.5
& \best{72.2} & 70.8 & \second{83.6}
& \second{48.8} & \best{61.5} & \second{43.5}
& \best{44.3} & \second{38.4} & \second{17.2}
& 23.5 & \second{45.5}
& \second{84.8} & 64.8\\

\textbf{GRSD (Ours)}
& \best{100.0} & \second{81.3} & 91.3 & \best{88.2}
& \second{70.8} & \best{85.7} & \best{86.7}
& \best{49.4} & \second{59.9} & 42.9
& \second{43.8} & \best{42.8} & \best{20.0}
& 16.2 & \best{46.2}
& \best{86.9} & \best{76.5}\\

\midrule
\rowcolor{backbonebg}
\multicolumn{18}{c}{\textit{Backbone: Qwen2.5-7B-Instruct}}\\
\midrule
Vanilla
& 45.2 & 27.3 & 13.8 & 8.3 & 4.8 & 0.0 & 18.0
& 25.5 & 50.2 & 29.0 & 28.6 & 29.0 & 10.1 & 62.5 & 32.6
& 2.3 & 0.0\\

Skill-Prompt
& 46.7 & 42.9 & 18.5 & 15.8 & 9.1 & 6.3 & 24.2
& 32.7 & 53.1 & 37.9 & 28.9 & 28.1 & 8.6 & 58.5 & 36.3
& 2.3 & 1.6\\

OPSD
& 71.4 & 40.0 & 25.0 & 40.0 & 16.7 & 19.2 & 35.9
& 1.2 & 1.6 & 1.7 & 0.7 & 2.4 & 0.1 & 0.4 & 1.4
& 5.5 & 2.3\\

GRPO
& \best{100.0} & 87.5 & 88.9 & 78.6
& \best{81.8} & \second{83.3} & 88.3
& 45.2 & 63.5 & \second{45.1} & 42.3
& 39.8 & 19.3 & 71.4 & 46.6
& 83.3 & 74.2\\

RLSD
& \best{100.0} & \second{90.9} & \second{92.6} & 75.0
& \second{81.0} & 68.2 & 84.8
& \second{47.4} & 63.4 & 44.5 & 44.4
& 36.4 & 19.5 & 71.0 & 45.8
& 87.4 & 76.6\\

Skill-SD
& \second{93.1} & 66.7 & \best{100.0} & 93.3
& 70.4 & 78.3 & 84.4
& 45.8 & 62.4 & 44.5 & 43.1
& 40.5 & 19.1 & 69.8 & 46.4
& 85.9 & 73.4\\

SDAR
& \best{100.0} & 66.7 & \best{100.0} & \best{94.1}
& 79.2 & \best{88.0} & \second{89.8}
& 47.0 & \second{63.7} & 43.6
& \best{45.4} & \second{44.1} & \best{20.9}
& \second{71.8} & \second{47.8}
& \second{89.6} & \second{78.1}\\

\textbf{GRSD (Ours)}
& \best{100.0} & \best{100.0} & \best{100.0}
& \second{93.4} & 66.7 & 82.4 & \best{92.2}
& \best{47.5} & \best{65.6} & \best{49.6}
& \second{44.5} & \best{47.1} & \second{20.7}
& \best{72.1} & \best{50.5}
& \best{91.6} & \best{82.8}\\

\bottomrule
\end{tabular}%
}
\caption[Main results across three backbones.]{
Main results across three backbones. We report success rate (\%) on ALFWorld,
accuracy (\%) on Search-based QA, and normalized Score/Success (\%) on WebShop.
\bestcap{Best} results are in bold, and
\secondcap{second-best} results are underlined.
}
\label{tab:main}
\end{table*}

\subsection{Overall Optimization Objective}
\label{sec:objective}

The full \textsc{GRSD} objective jointly improves task execution and the
policy's ability to reflect on its verified experience:
\begin{equation}
\label{eq:total_loss}
    \mathcal{L}(\theta)
    =
    \mathcal{L}_{\mathrm{task}}(\theta)
    +
    \alpha\,\mathcal{L}_{\mathrm{ref}}(\theta),
\end{equation}
where $\alpha$ controls the contribution of reflection learning.
The task loss follows the clipped GRPO objective, but replaces the trajectory-level advantage $A_i$ with the turn-level modulated advantage $\widehat{A}_{i,t}$:
\begin{equation}
\label{eq:task_loss}
\begin{aligned}
    \mathcal{L}_{\mathrm{task}}(\theta)
    &=
    -\mathbb{E}
    \Bigg[
    \frac{1}{G}
    \sum_{i=1}^{G}
    \frac{1}{N_i}
    \sum_{t=1}^{T_i}
    \sum_{k=1}^{L_{i,t}}
    \min
    \Big(
    \rho_{i,t,k}\widehat{A}_{i,t},\\
    &\qquad\qquad
    \mathrm{clip}(\rho_{i,t,k},1-\varepsilon,1+\varepsilon)
    \widehat{A}_{i,t}
    \Big)
    \Bigg].
\end{aligned}
\end{equation}
Here, $\widehat{A}_{i,t}$ is shared by the action tokens within turn $t$ but different across turns of the same trajectory.

The reflection loss optimizes only the reflection tokens sampled in Stage~A.
It uses the same GRPO surrogate as Eq.~(\ref{eq:grpo}), applied to the
reflection sequence $s_i$ with its independently normalized advantage
$A^{\mathrm{ref}}_i$:
\begin{equation}
\label{eq:reflection_loss}
    \mathcal{L}_{\mathrm{ref}}(\theta)
    =
    \mathcal{L}_{\mathrm{GRPO}}^{\mathrm{ref}}
    (\theta;A^{\mathrm{ref}}).
\end{equation}

The two objectives supervise disjoint token sets:
$\mathcal{L}_{\mathrm{task}}$ updates task-trajectory tokens, whereas
$\mathcal{L}_{\mathrm{ref}}$ trains reflection tokens $s_i$.
The group-level guidance construction and teacher--student evaluations are
treated as stop-gradient, while the external judge serves solely to provide
reflection rewards for $\mathcal{L}_{\mathrm{ref}}$.
Within $\mathcal{L}_{\mathrm{task}}$, privileged guidance affects optimization
only through detached credit modulation and is not used as a generation target.

\begin{figure*}[t]
    \centering
    \includegraphics[width=\linewidth]{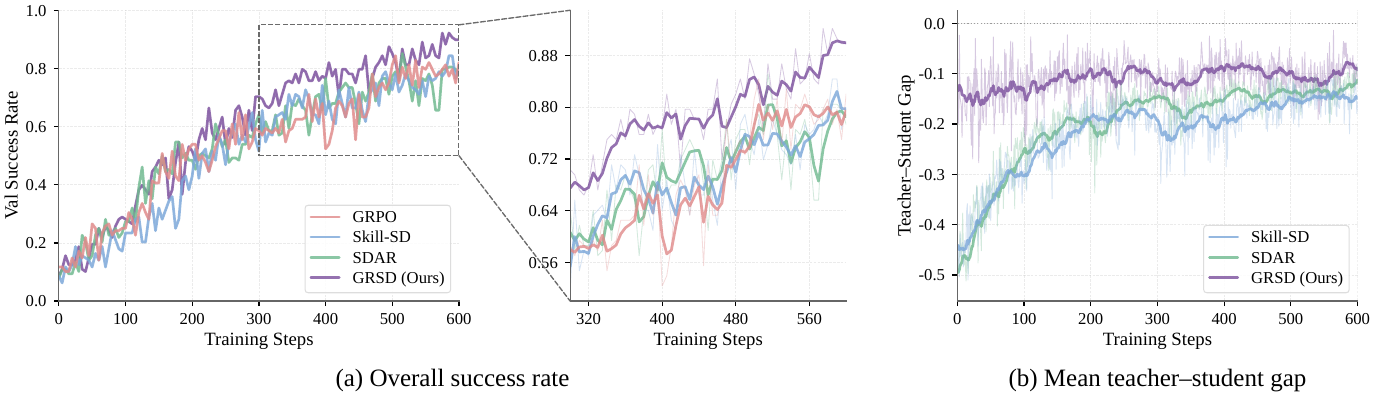}
    \caption{Training dynamics on ALFWorld using Qwen3-1.7B. 
    (a) Validation success rate over training steps. 
    (b) Mean teacher--student gap $\bar{\Delta}$ under the two conditioning contexts.}
    \label{fig:dynamics}
\end{figure*}

\section{Experiment}

\subsection{Experimental Setup}
\label{sec:exp_setup}

\paragraph{Benchmarks.}
Following recent work on self-distillation, we
evaluate on three representative long-horizon environments spanning embodied
reasoning, search-augmented question answering, and web navigation.
ALFWorld~\citep{shridhar2020alfworld} is a text-based embodied environment
aligned with the ALFRED benchmark. We report the success rate across its six
task types: Pick, Look, Clean, Heat, Cool, and Pick2.
Search-based QA~\citep{jin2025search} follows the Search-R1 setup, where the
agent gathers external evidence through search before producing an answer.
We train on two in-domain datasets, NQ~\citep{kwiatkowski2019natural} and
HotpotQA~\citep{yang2018hotpotqa}, and evaluate out-of-domain generalization
on five additional datasets: TriviaQA~\citep{joshi2017triviaqa},
PopQA~\citep{mallen2023not}, 2Wiki~\citep{ho2020constructing},
MuSiQue~\citep{trivedi2022musique}, and
Bamboogle~\citep{press2023measuring}.
WebShop~\citep{yao2022webshop} is an interactive online-shopping environment
evaluated on 128 held-out tasks. 
Details are provided in the supplementary material.

\paragraph{Baselines.}
We compare \textsc{GRSD} against three groups of methods.
\emph{(1) Training-free}: Vanilla, the instruction-tuned model without
task-specific training, and Skill-Prompt, which augments the task prompt with
retrieved skills at inference.
\emph{(2) Outcome-based RL and general self-distillation}: 
GRPO~\citep{shao2024deepseekmath}, which optimizes outcome-based verifiable
rewards, and OPSD~\citep{zhao2026self} and RLSD~\citep{yang2026self}, which
add self-distillation without explicitly constructing skills.
\emph{(3) Skill-based self-distillation}: 
Skill-SD~\citep{wang2026skill} and SDAR~\citep{lu2026self}, which incorporate
skill-conditioned privileged guidance into policy optimization.

\paragraph{Models and Metrics.}
To assess robustness across model families and scales, we evaluate
Qwen3-1.7B, Qwen2.5-3B-Instruct, and Qwen2.5-7B-Instruct. We report success
rate (\%) on ALFWorld, answer accuracy (\%) on Search-based QA, and normalized
score and success rate (\%) on WebShop. The reported averages are computed over all evaluation instances, weighted
by subset size.

\paragraph{Implementation Details.}
All methods utilize veRL with FSDP and vLLM, and are trained on one node with eight NVIDIA H100 GPUs under identical backbones, environment wrappers, rollout budgets, and schedules. We adopt AdamW with a learning rate of $1\times10^{-6}$, a GRPO group size of $G=8$, a KL coefficient of $0.01$, and a mini-batch size of $256$ turns. For ALFWorld, we follow the GiGPO split~\citep{feng2026group} with $16$ tasks per batch, prompt/response limits of $2{,}048/512$ tokens, and at most $50$ environment steps. For Search-based QA, we follow Search-R1~\citep{jin2025search}, use E5~\citep{wang2022text} for retrieval, and train on NQ and HotpotQA with $128$ tasks per batch, $4{,}096/512$ tokens, and at most $4$ environment steps. For WebShop, we use $1{,}000$ training tasks with $16$ per batch, $4{,}096/512$ tokens, and at most $15$ environment steps. For \textsc{GRSD}, we set $\lambda=0.5$, $\widehat{G}=0.2$, and $\alpha=0.01$. During training, a fixed DeepSeek-V4-Flash judge with temperature $0$ scores each trajectory reflection on a four-point scale. Further configurations and prompt templates are provided in the supplementary material.

\subsection{Main Results}
\label{sec:main_results}
Table~\ref{tab:main} reports the main results across all three environments and backbones. We discuss the key observations below.

\paragraph{\textsc{GRSD} achieves the strongest aggregate performance.}
Across all three backbones, \textsc{GRSD} substantially surpasses Vanilla and
Skill-Prompt, demonstrating benefits beyond inference-time skill augmentation.
Averaged across backbones, it outperforms GRPO by $5.5\%$ on ALFWorld and
$2.7\%$ on Search-based QA, while improving WebShop score and success by
$4.8\%$ and $7.2\%$, respectively. It also achieves the best aggregate result
among GRPO, OPSD, and RLSD for every environment and backbone.

\paragraph{\textsc{GRSD} further advances skill-based self-distillation.}
Compared with the stronger of Skill-SD and SDAR for each backbone,
\textsc{GRSD} improves ALFWorld overall success by $4.2\%$ and WebShop success
by $5.5\%$ on average, while retaining the best cross-backbone averages on
Search-based QA and WebShop score. Although gains are not uniform on every
individual subset, \textsc{GRSD} attains the strongest overall performance
under all three backbones. These results support the benefit of contrasting
policy-native reflections across successful and failed rollouts.

\subsection{Training Dynamics}
\label{sec:training_dynamics}

Figure~\ref{fig:dynamics} compares the training dynamics of \textsc{GRSD} with
representative baselines on ALFWorld. All methods improve rapidly early in
training, whereas \textsc{GRSD} continues to improve after the baselines begin
to plateau and ultimately reaches the highest validation success rate. The
enlarged window from Step 300 onward shows that this advantage persists through
the second half of training rather than arising from a transient peak.
\textsc{GRSD} also maintains a smaller absolute mean teacher--student gap than
Skill-SD and SDAR across nearly all training steps. This indicates that its
privileged context induces a smaller shift from the policy's native action
likelihoods, a trend consistent with more compatible guidance and smoother
late-stage optimization.

\begin{table}[t]
\centering
\resizebox{\columnwidth}{!}{%
\begin{tabular}{l|c|c|cc}
\toprule
\multirow{2}{*}{Variant}
& ALFWorld
& SearchQA
& \multicolumn{2}{c}{WebShop}\\
\cmidrule(lr){2-2}
\cmidrule(lr){3-3}
\cmidrule(lr){4-5}
& Avg & Avg & Score & Succ\\
\midrule
\textbf{GRSD (full)}
& \textbf{92.2} & \textbf{43.8} & \textbf{84.5} & \textbf{66.4}\\
\midrule
\textbf{\textit{Guidance construction}} & & & &\\
\quad w/o group-reflective guidance
& 85.2 & 42.4 & 79.0 & 59.4\\
\quad w/o group aggregation
& 89.8 & 42.7 & 83.2 & 65.6\\
\quad w/o failed reflections $\mathcal{S}^{-}_x$
& 88.3 & 43.1 & 82.2 & 62.5\\
\midrule
\textbf{\textit{Advantage modulation}} & & & &\\
\quad w/ token-level
& 89.1 & 43.4 & 83.0 & 64.1\\
\quad w/ trajectory-level
& 85.9 & 42.9 & 79.6 & 61.0\\
\bottomrule
\end{tabular}%
}
\caption{Ablation of group-reflective guidance construction and advantage modulation in
\textsc{GRSD} using Qwen3-1.7B.}
\label{tab:ablation}
\end{table}

\subsection{Ablation Studies}
\label{sec:ablation}

% We evaluate the key components of \textsc{GRSD} across all three environments in Table~\ref{tab:ablation}, with detailed analyses on ALFWorld in Figures~\ref{fig:dynamics_ablation} and \ref{fig:hyper_ablation}.

\paragraph{Effectiveness of group-reflective guidance.}
We evaluate three guidance variants:
\emph{w/o group-reflective guidance} replaces $z_x$ with a standard reference
solution, \emph{w/o group aggregation} bypasses Stage~B and uses
unaggregated trajectory reflections, and \emph{w/o failed reflections}
constructs $z_x$ using only $\mathcal{S}^{+}_x$. As shown in
Table~\ref{tab:ablation}, all three variants degrade performance, confirming
their complementary roles. Replacing group-reflective guidance causes
substantial drops of $7.0\%$ on ALFWorld and WebShop success and $5.5\%$ on
WebShop score, demonstrating the value of guidance derived from the current
on-policy rollout group. Bypassing group aggregation reduces ALFWorld
performance by $2.4\%$ and degrades the other metrics, supporting Stage~B in
consolidating patterns across trajectories. Excluding failed reflections
leads to $3.9\%$ drops on ALFWorld and WebShop success and a $2.3\%$ decrease
in WebShop score, showing that failure modes provide essential information
beyond successful behavior.

\begin{figure}[t]
\centering

\begin{minipage}[t]{0.49\linewidth}
    \centering
    \includegraphics[width=\linewidth]{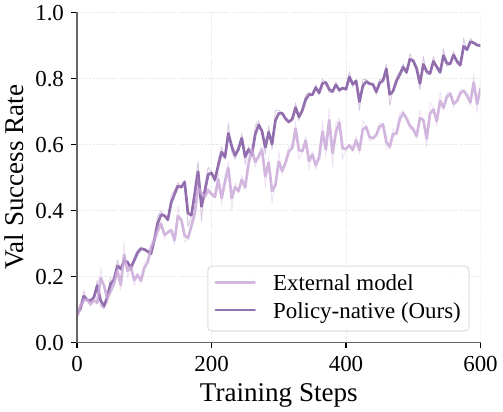}
    \vspace{-0.6em}

    {\small (a) Overall success rate\par}
\end{minipage}
\hfill
\begin{minipage}[t]{0.49\linewidth}
    \centering
    \includegraphics[width=\linewidth]{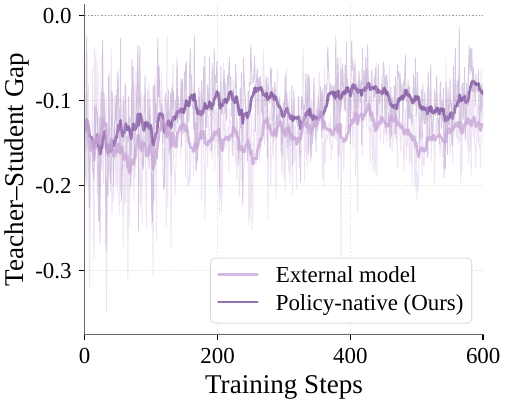}
    \vspace{-0.6em}

    {\small (b) Mean teacher--student gap\par}
\end{minipage}

\vspace{-0.5em}
\caption{Comparison of the complete policy-native reflection pipeline with
external-model reflection generation on ALFWorld using Qwen3-1.7B.
(a) Overall success rate. 
(b) Mean teacher--student gap.}
\label{fig:dynamics_ablation}
\end{figure}

\begin{figure}[t]
\centering

\begin{minipage}[t]{0.49\linewidth}
    \centering
    \includegraphics[width=\linewidth]{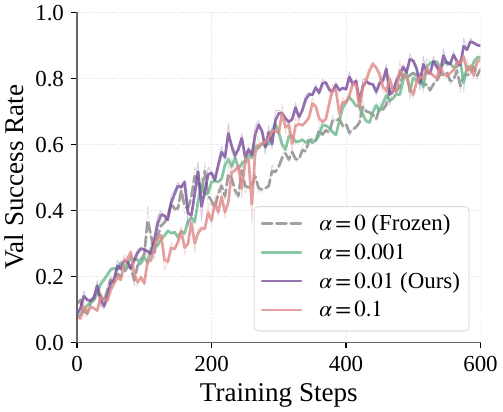}
    \vspace{-0.6em}

    {\small (a) Reflection-loss weight $\alpha$\par}
\end{minipage}
\hfill
\begin{minipage}[t]{0.49\linewidth}
    \centering
    \includegraphics[width=\linewidth]{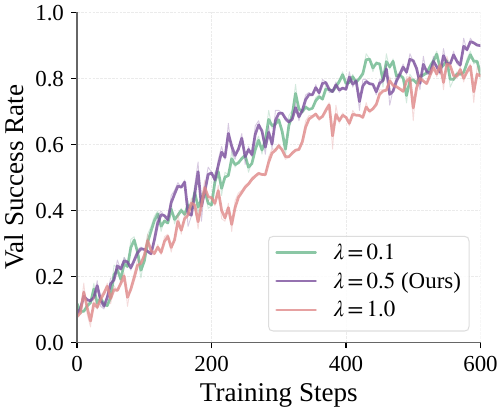}
    \vspace{-0.6em}

    {\small (b) Self-distillation strength $\lambda$\par}
\end{minipage}

\vspace{-0.5em}
\caption{Hyperparameter sensitivity analysis on ALFWorld with Qwen3-1.7B. 
(a) Effect of the self-reflection loss weight $\alpha$. 
(b) Effect of the self-distillation strength $\lambda$.}
\label{fig:hyper_ablation}
\end{figure}

\begin{figure}[t]
\centering
\includegraphics[width=0.98\linewidth]{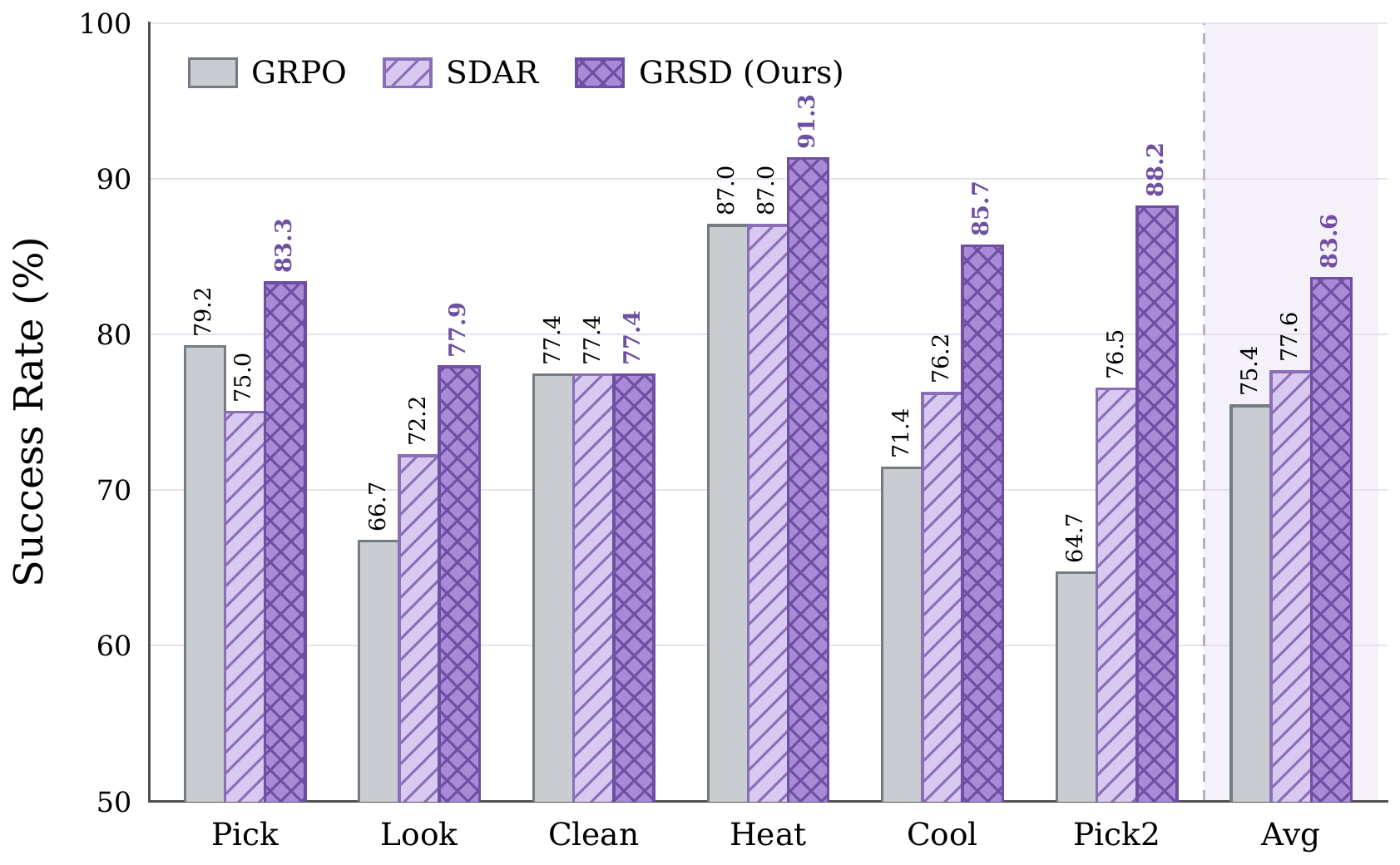}
\caption{Generalization performance on the ALFWorld Unseen split. \textsc{GRSD} achieves
the highest overall success rate, with particularly large gains on Pick2 and
Cool.}
\label{fig:generalization_unseen}
\end{figure}

\paragraph{Turn-level modulation provides the best granularity.}
In Table~\ref{tab:ablation}, we compare token-level modulation over action tokens with
trajectory-level modulation using one shared rollout factor. Both degrade all
metrics: token-level modulation reduces ALFWorld and WebShop success by
$3.1\%$ and $2.3\%$, while trajectory-level modulation causes drops of
$6.3\%$, $0.9\%$, and $5.4\%$ on ALFWorld, SearchQA, and WebShop success, respectively. Turn-level modulation thus better
aligns credit with interaction decisions while avoiding token-level noise and
coarse trajectory-level attribution.

\paragraph{Policy-native reflections provide more compatible guidance.} Figure~\ref{fig:dynamics_ablation}(a) shows that policy-native reflection generation achieves a higher success rate than its external-model counterpart, with a clear advantage in the second half of training. It also maintains a lower magnitude of the mean teacher--student gap throughout training in Figure~\ref{fig:dynamics_ablation}(b). These results suggest that reflections generated by the optimized policy better match its capability boundary and therefore provide guidance that can be more effectively internalized.

\paragraph{Effect of the reflection-loss weight $\alpha$.}
Figure~\ref{fig:hyper_ablation}(a) compares
$\alpha\in\{0,0.001,0.01,0.1\}$. The moderate setting $\alpha=0.01$ achieves
the strongest late-stage performance, indicating that learning to reflect on
the policy's behavior gradually improves subsequent task behavior. Freezing reflection-specific learning at $\alpha=0$ limits later benefits, while $\alpha=0.001$ provides insufficient supervision to effectively shape reflective behavior. 
In contrast, $\alpha=0.1$ causes the auxiliary reflection objective to compete excessively with task optimization, diverting learning capacity from task completion.

\paragraph{Effect of the distillation strength $\lambda$.} Figure~\ref{fig:hyper_ablation}(b) compares $\lambda\in\{0.1,0.5,1.0\}$. A small $\lambda$ leaves turn-level advantages close to the original trajectory-level signal, limiting decision-level differentiation, whereas an overly large $\lambda$ excessively reshapes them based on potentially noisy teacher--student discrepancies. The best performance at $\lambda=0.5$ shows that moderate modulation introduces informative turn-level credit while preserving the verifier-grounded trajectory signal.

\subsection{Generalization on ALFWorld Unseen}
\label{sec:zero-shot}
Figure~\ref{fig:generalization_unseen} evaluates generalization on the
ALFWorld Unseen split. \textsc{GRSD} achieves an average success rate of
$83.6\%$, outperforming GRPO and SDAR by $8.2\%$ and $6.0\%$, respectively.
The largest gains occur on Pick2 and Cool, where \textsc{GRSD} exceeds GRPO by
$23.5\%$ and $14.3\%$, and SDAR by $11.7\%$ and $9.5\%$, respectively.
By contrasting successful and failed policy-native reflections, \textsc{GRSD}
captures success-critical behaviors and recurring failure modes while reducing
trajectory-specific details. Internalizing this capability-compatible guidance
encourages more transferable decision principles, leading to stronger
performance on unseen tasks.

\section{Conclusion}
We presented \textsc{GRSD}, a group-reflective self-distillation framework for agentic reinforcement learning with sparse outcome rewards. The policy reflects on verified on-policy trajectories, while a stop-gradient snapshot contrasts successful and failed reflections to form group-level privileged guidance. This policy-native construction yields capability-aligned and outcome-discriminative guidance without external content generation. During training, a self-teacher uses it to redistribute trajectory-level advantages across turns without reversing the verifier-determined direction, adding no inference overhead. Across environments and model scales, \textsc{GRSD} achieves the strongest aggregate performance, stronger unseen-task generalization, and sustained late-stage improvements. By extracting and internalizing guidance from its own verified experience, \textsc{GRSD} establishes an experience-driven loop for agentic self-evolution.

\bibliography{aaai2027}

\clearpage
\appendix

\clearpage
\twocolumn

\section{Theoretical Analysis of \textsc{GRSD}}
\label{app:theory}

We analyze how \textsc{GRSD} converts group-reflective guidance into a bounded
turn-wise reweighting of the GRPO policy surrogate. For a prompt $x$, let
$\{\tau_i\}_{i=1}^{G}$ denote a rollout group, $R_i$ the verifier reward, and
$b_i\in\{0,1\}$ the binary outcome label used to partition successful and failed
reflections. For $\epsilon_{\mathrm{adv}}>0$, the trajectory-level GRPO
advantage is
\begin{equation}
    A_i
    =
    \frac{R_i-\bar R}
    {
    \sqrt{
    \frac{1}{G}\sum_{j=1}^{G}(R_j-\bar R)^2
    +\epsilon_{\mathrm{adv}}
    }
    },
    \quad
    \bar R
    =
    \frac{1}{G}\sum_{j=1}^{G}R_j.
\label{eq:app_advantage}
\end{equation}

The group-level guidance, teacher--student likelihood scores, and turn weights
considered below are computed using the frozen policy snapshot
$\pi_{\bar\theta}$, where
$\bar\theta=\operatorname{sg}(\theta_{\mathrm{old}})$, and are treated as
constants in the task-policy surrogate. The reflections used to construct the
guidance are sampled from the same snapshot; their separate learning path
through the reflection objective is discussed below.

\subsection{Group-Reflective Contrast}

Let $s_i$ be the reflection generated from trajectory $\tau_i$, and let
$h(s_i)\in\mathbb R^d$ denote a fixed representation of that reflection. To
characterize the benefit of aggregating reflections within a rollout group, we
consider the idealized model
\begin{equation}
    h(s_i)
    =
    \mu_x^{b_i}+\xi_i,
\label{eq:app_reflection_model}
\end{equation}
where
\begin{equation}
    \mathbb E[\xi_i\mid x,\mathbf b]=0,
    \qquad
    \operatorname{Cov}(\xi_i\mid x,\mathbf b)
    =
    \Sigma_x^{b_i},
\end{equation}
and the noise terms are conditionally pairwise uncorrelated given the prompt and
outcome labels.

For a mixed-outcome group, define
\begin{equation}
\begin{aligned}
    I_x^+ &= \{i:b_i=1\},
    & n_+ &= |I_x^+|,\\
    I_x^- &= \{i:b_i=0\},
    & n_- &= |I_x^-|.
\end{aligned}
\end{equation}
The corresponding group-level contrast estimator is
\begin{equation}
    \widehat d_x
    =
    \frac{1}{n_+}
    \sum_{i\in I_x^+}h(s_i)
    -
    \frac{1}{n_-}
    \sum_{i\in I_x^-}h(s_i).
\label{eq:app_group_contrast}
\end{equation}

\paragraph{Proposition 1 (Variance reduction in group-level contrast).}
Under the model above,
\begin{equation}
    \mathbb E[\widehat d_x\mid x,\mathbf b]
    =
    \mu_x^1-\mu_x^0,
\label{eq:app_contrast_expectation}
\end{equation}
and
\begin{equation}
\begin{aligned}
    \mathbb E\!\left[
    \left\|
    \widehat d_x-(\mu_x^1-\mu_x^0)
    \right\|_2^2
    \mid x,\mathbf b
    \right]
    =
    \operatorname{tr}
    \left(
    \frac{\Sigma_x^1}{n_+}
    +
    \frac{\Sigma_x^0}{n_-}
    \right).
\end{aligned}
\label{eq:app_contrast_variance}
\end{equation}

\emph{Proof.}
By linearity of expectation, the means of the successful and failed reflection
sets are $\mu_x^1$ and $\mu_x^0$, respectively, which proves
Eq.~\eqref{eq:app_contrast_expectation}. Conditional uncorrelatedness gives
\begin{equation}
    \operatorname{Cov}
    (\widehat d_x\mid x,\mathbf b)
    =
    \frac{\Sigma_x^1}{n_+}
    +
    \frac{\Sigma_x^0}{n_-}.
\end{equation}
Because $\widehat d_x$ is conditionally unbiased, its conditional mean-squared
error equals the trace of this covariance.

A contrast constructed from one successful and one failed reflection has
conditional mean-squared error
$\operatorname{tr}(\Sigma_x^1+\Sigma_x^0)$. Group aggregation therefore reduces
trajectory-specific variation whenever either subset contains multiple
non-degenerate observations. This proposition characterizes the evidence
available to Stage~B under an idealized representation model. The actual method
aggregates natural-language reflections through the frozen policy and does not
explicitly compute $\widehat d_x$; hence, the result does not imply that the
generated guidance exactly recovers $\mu_x^1-\mu_x^0$.

The conditional-uncorrelatedness assumption excludes dependencies induced by
shared prompts, policy parameters, or decoding behavior. If the reflection
noise is conditionally correlated, the covariance of $\widehat d_x$ additionally
contains cross-covariance terms, and the inverse-subset-size reduction in
Eq.~\eqref{eq:app_contrast_variance} need not hold. Proposition~1 should
therefore be interpreted as characterizing the benefit of aggregating
sufficiently diverse reflective evidence, rather than as an unconditional
guarantee for the natural-language guidance generator.

\subsection{Turn-Level Likelihood Signal}

For turn $t$ of trajectory $i$, let $H_{i,t}$ denote the interaction history
available before that turn and let
\begin{equation}
    y_{i,t}
    =
    (y_{i,t,1},\ldots,y_{i,t,L_{i,t}})
\end{equation}
denote its sampled action tokens. The plain student context and the
guidance-augmented teacher context are
\begin{equation}
\begin{aligned}
    c_{i,t}^{\mathrm{stu}}
    &=
    \mathcal C(x,H_{i,t}),\\
    c_{i,t}^{\mathrm{tea}}
    &=
    \mathcal C(x,z_x,H_{i,t}),
\end{aligned}
\label{eq:app_contexts}
\end{equation}
where $z_x$ is the group-level privileged guidance. The same frozen policy
evaluates the same sampled action in both contexts. The likelihood difference
therefore isolates the effect of adding $z_x$, rather than differences in model
parameters or capacity.

The mean sampled-token log likelihoods are
\begin{equation}
\begin{aligned}
    \ell_{i,t}^{\mathrm{tea}}
    &=
    \frac{1}{L_{i,t}}
    \sum_{k=1}^{L_{i,t}}
    \log\pi_{\bar\theta}
    \left(
    y_{i,t,k}
    \mid
    y_{i,t,<k},
    c_{i,t}^{\mathrm{tea}}
    \right),\\
    \ell_{i,t}^{\mathrm{stu}}
    &=
    \frac{1}{L_{i,t}}
    \sum_{k=1}^{L_{i,t}}
    \log\pi_{\bar\theta}
    \left(
    y_{i,t,k}
    \mid
    y_{i,t,<k},
    c_{i,t}^{\mathrm{stu}}
    \right).
\end{aligned}
\label{eq:app_turn_likelihoods}
\end{equation}
Their difference is
\begin{equation}
    \Delta_{i,t}
    =
    \ell_{i,t}^{\mathrm{tea}}
    -
    \ell_{i,t}^{\mathrm{stu}},
    \qquad
    q_{i,t}
    =
    \operatorname{sign}(A_i)\Delta_{i,t}.
\label{eq:app_aligned_gap}
\end{equation}

Since the likelihood is averaged over action tokens,
$\exp(\Delta_{i,t})$ corresponds to the geometric-mean likelihood ratio assigned
to the sampled action under the guidance-augmented and plain contexts. Thus,
$q_{i,t}>0$ indicates that the guidance-induced likelihood change agrees with
the reward-relative update direction encoded by $A_i$: it increases the
likelihood of a positive-advantage action or decreases the likelihood of a
negative-advantage action.

The outcome label $b_i$ and verifier reward $R_i$ play different roles.
Specifically, $b_i$ determines whether a reflection enters the successful or
failed subset, whereas the sign of $A_i$ is determined by whether $R_i$ lies
above or below the group mean. These directions coincide exactly in the
binary-reward case $R_i=b_i\in\{0,1\}$, but need not coincide when the
environment provides continuous rewards.

Let $\mathcal T_x$ be the set of valid action turns in the rollout group.
Assuming $\mathcal T_x\neq\varnothing$ and $\epsilon_{\mathrm{gap}}>0$, the
aligned likelihood gaps are converted into normalized and clipped modulation
scores:
\begin{equation}
\begin{aligned}
    u_{i,t}
    &=
    \exp(q_{i,t})-1,\\
    D_x
    &=
    \frac{1}{|\mathcal T_x|}
    \sum_{(j,t')\in\mathcal T_x}
    |u_{j,t'}|
    +
    \epsilon_{\mathrm{gap}},\\
    \widehat q_{i,t}
    &=
    \operatorname{clip}
    \left(
    \frac{u_{i,t}}{D_x},
    -\widehat G,
    \widehat G
    \right).
\end{aligned}
\label{eq:app_gap_definition}
\end{equation}
The normalization reduces prompt-dependent scale variation, while clipping
limits the influence of extreme likelihood changes. The exponential
transformation is strictly increasing and zero-centered; division by the
positive normalizer and symmetric clipping with $\widehat G>0$ preserve its
sign. Therefore,
\begin{equation}
    \operatorname{sign}(\widehat q_{i,t})
    =
    \operatorname{sign}(q_{i,t}),
    \qquad
    |\widehat q_{i,t}|
    \leq
    \widehat G.
\label{eq:app_gap_bound}
\end{equation}

If the rollout group does not contain both successful and failed trajectories,
or if valid guidance or valid action turns are unavailable, we set
$\widehat q_{i,t}=0$. In these cases, \textsc{GRSD} retains the original GRPO
advantage.

\subsection{Bounded Advantage Modulation}

Let
\begin{equation}
    m_x
    =
    \mathbf 1
    \left[
    |I_x^+|>0
    \ \wedge\
    |I_x^-|>0
    \right]
\end{equation}
denote the mixed-group mask. The detached turn weight and modulated advantage
are
\begin{equation}
\begin{aligned}
    w_{i,t}
    &=
    1+\lambda m_x\widehat q_{i,t},\\
    \widehat A_{i,t}
    &=
    A_iw_{i,t}.
\end{aligned}
\label{eq:app_turn_weight}
\end{equation}

\paragraph{Proposition 2 (Bounded positive modulation).}
Suppose $\lambda\geq0$ and $\widehat G>0$, and let
$\delta=\lambda\widehat G$. If $\delta<1$, then
\begin{equation}
    1-\delta
    \leq
    w_{i,t}
    \leq
    1+\delta,
\label{eq:app_weight_bound}
\end{equation}
and
\begin{equation}
    |\widehat A_{i,t}-A_i|
    \leq
    \delta|A_i|.
\label{eq:app_modulation_bound}
\end{equation}
Moreover, for every $A_i\neq0$,
\begin{equation}
    \operatorname{sign}(\widehat A_{i,t})
    =
    \operatorname{sign}(A_i).
\label{eq:app_sign_preservation}
\end{equation}

\emph{Proof.}
From Eq.~\eqref{eq:app_gap_bound} and $m_x\in\{0,1\}$,
\begin{equation}
    |\lambda m_x\widehat q_{i,t}|
    \leq
    \lambda\widehat G
    =
    \delta.
\end{equation}
This gives Eq.~\eqref{eq:app_weight_bound}. Furthermore,
\begin{equation}
\begin{aligned}
    |\widehat A_{i,t}-A_i|
    &=
    |A_i||w_{i,t}-1|\leq
    \delta|A_i|.
\end{aligned}
\end{equation}
Since $\delta<1$, $w_{i,t}>0$, and multiplication by $w_{i,t}$ preserves the
sign of every nonzero advantage.

Consequently, aligned likelihood evidence increases the magnitude of the
corresponding advantage, whereas conflicting evidence decreases it. The
guidance can refine the strength of the verifier-derived signal but cannot
replace it with the opposite direction. Symbolically, the relative perturbation
is at most $\delta=\lambda\widehat G$, and
$w_{i,t}\in[1-\delta,1+\delta]$. The case $\delta=0$ recovers the original GRPO
advantage, while any $\delta<1$ keeps every turn weight strictly positive.

\subsection{Positive Reweighting of the GRPO Surrogate}

For token $k$ in turn $t$ of trajectory $i$, define the importance ratio under
the plain task context:
\begin{equation}
    \rho_{i,t,k}(\theta)
    =
    \frac{
    \pi_\theta
    \left(
    y_{i,t,k}
    \mid
    y_{i,t,<k},
    c_{i,t}^{\mathrm{stu}}
    \right)
    }{
    \pi_{\theta_{\mathrm{old}}}
    \left(
    y_{i,t,k}
    \mid
    y_{i,t,<k},
    c_{i,t}^{\mathrm{stu}}
    \right)
    }.
\label{eq:app_policy_ratio}
\end{equation}
For clipping radius $\varepsilon\in[0,1)$, let
\begin{equation}
    \phi_\varepsilon(\rho,A)
    =
    \min
    \left(
    \rho A,\,
    \operatorname{clip}
    (\rho,1-\varepsilon,1+\varepsilon)A
    \right)
\label{eq:app_clipped_term}
\end{equation}
denote one clipped GRPO surrogate term.

\paragraph{Proposition 3 (Positive reweighting equivalence).}
For any $w\geq0$,
\begin{equation}
    \phi_\varepsilon(\rho,wA)
    =
    w\phi_\varepsilon(\rho,A).
\label{eq:app_positive_homogeneity}
\end{equation}

\emph{Proof.}
Both arguments of the minimum in Eq.~\eqref{eq:app_clipped_term} are multiplied
by the same nonnegative scalar:
\begin{equation}
\begin{aligned}
    \phi_\varepsilon(\rho,wA)
    &=
    \min
    \left(
    w\rho A,\,
    w\operatorname{clip}
    (\rho,1-\varepsilon,1+\varepsilon)A
    \right)\\
    &=
    w\phi_\varepsilon(\rho,A).
\end{aligned}
\end{equation}

Let $\omega_{i,t,k}\geq0$ denote the coefficient induced by the chosen averaging
scheme. Applying Proposition~3 with $w=w_{i,t}$ gives
\begin{equation}
\begin{aligned}
    \mathcal L_{\mathrm{surr}}^{\mathrm{GRSD}}(\theta)
    =
    -\mathbb E
    \left[
    \sum_{i,t,k}
    \omega_{i,t,k}w_{i,t}
    \phi_\varepsilon
    \left(
    \rho_{i,t,k}(\theta),
    A_i
    \right)
    \right].
\end{aligned}
\label{eq:app_weighted_surrogate}
\end{equation}
Therefore, the policy-surrogate component of \textsc{GRSD} is exactly a
detached, positively reweighted clipped GRPO surrogate. This identity applies
to the policy surrogate itself; independently computed KL penalties and the
reflection objective are not reweighted.

Define the per-token surrogate gradient contributions as
\begin{equation}
\begin{aligned}
    g_{i,t,k}^{\mathrm{GRPO}}
    &:=
    -\nabla_\theta
    \left[
    \omega_{i,t,k}
    \phi_\varepsilon
    \left(
    \rho_{i,t,k}(\theta),A_i
    \right)
    \right],\\
    g_{i,t,k}^{\mathrm{GRSD}}
    &:=
    -\nabla_\theta
    \left[
    \omega_{i,t,k}w_{i,t}
    \phi_\varepsilon
    \left(
    \rho_{i,t,k}(\theta),A_i
    \right)
    \right].
\end{aligned}
\label{eq:app_gradient_definitions}
\end{equation}
Because $w_{i,t}$ is detached, these contributions satisfy, almost everywhere,
\begin{equation}
    g_{i,t,k}^{\mathrm{GRSD}}
    =
    w_{i,t}g_{i,t,k}^{\mathrm{GRPO}}.
\label{eq:app_pointwise_gradient}
\end{equation}
Together with Eq.~\eqref{eq:app_weight_bound}, this gives
\begin{equation}
    \left\|
    g_{i,t,k}^{\mathrm{GRSD}}
    -
    g_{i,t,k}^{\mathrm{GRPO}}
    \right\|
    \leq
    \delta
    \left\|
    g_{i,t,k}^{\mathrm{GRPO}}
    \right\|.
\label{eq:app_gradient_bound}
\end{equation}
Thus, each individual surrogate contribution undergoes a bounded positive
rescaling. For a realized minibatch, let
$g^{\mathrm{GRPO}}=\sum_{i,t,k}g_{i,t,k}^{\mathrm{GRPO}}$ and
$g^{\mathrm{GRSD}}=\sum_{i,t,k}g_{i,t,k}^{\mathrm{GRSD}}$. The triangle
inequality further yields
\begin{equation}
    \left\|
    g^{\mathrm{GRSD}}-g^{\mathrm{GRPO}}
    \right\|
    \leq
    \delta
    \sum_{i,t,k}
    \left\|
    g_{i,t,k}^{\mathrm{GRPO}}
    \right\|.
\label{eq:app_aggregate_gradient_bound}
\end{equation}
This is an absolute perturbation bound relative to the sum of individual
gradient norms, not a relative bound in terms of
$\|g^{\mathrm{GRPO}}\|$. In particular, positive per-term reweighting does not
imply that the aggregate GRSD and GRPO gradients are collinear, because
different sample gradients may point in different directions and partially
cancel.

\subsection{Gradient Routing and Reflection Learning}

The stop-gradient construction separates the estimation of turn weights from
the task-policy update. Since all quantities used to construct $w_{i,t}$ are
computed by the frozen snapshot and detached,
\begin{equation}
    \nabla_\theta w_{i,t}=0.
\end{equation}
The task loss therefore optimizes the plain-context action likelihood using
fixed turn-level weights, without allowing the task gradient to directly modify
the mechanism producing those weights.

Stage~A remains trainable through the separate reflection objective. Reflection
sequences are sampled from the frozen snapshot and their judge-derived
advantages are treated as fixed, but their log likelihoods under the current
policy still depend on $\theta$. Let $\mathcal L_{\mathrm{ref}}$ denote the
resulting GRPO-style reflection loss. The complete objective is
\begin{equation}
    \mathcal L
    =
    \mathcal L_{\mathrm{surr}}^{\mathrm{GRSD}}
    +
    \beta\mathcal L_{\mathrm{KL}}
    +
    \alpha\mathcal L_{\mathrm{ref}}.
\label{eq:app_joint_objective}
\end{equation}
Accordingly,
\begin{equation}
\begin{aligned}
    \nabla_\theta\mathcal L
    =
    {}&
    \nabla_\theta
    \mathcal L_{\mathrm{surr}}^{\mathrm{GRSD}}
    +
    \beta\nabla_\theta\mathcal L_{\mathrm{KL}}+
    \alpha\nabla_\theta\mathcal L_{\mathrm{ref}}.
\end{aligned}
\label{eq:app_joint_gradient}
\end{equation}

The task and reflection objectives supervise different sampled token sets but
share the same policy parameters. The reflection objective provides the
policy-native reflection generator with judge feedback, which can make later
policy snapshots produce more informative reflective evidence. In contrast,
the Stage~B guidance and turn weights receive no direct task-loss gradient
within the current update.

\paragraph{Scope of the analysis.}
Under the representation assumptions of Proposition~1, aggregating successful
and failed reflections can reduce trajectory-specific variation in an
idealized contrast statistic. Independently of that statistical model,
Propositions~2 and~3 establish that the resulting detached likelihood signal,
once normalized and clipped as specified, induces a bounded positive
reweighting of the clipped GRPO policy surrogate.

These conclusions concern the optimization mechanism rather than policy
performance. In particular, $z_x$ may depend on complete trajectories,
including the trajectory whose turns are being scored, so
$\Delta_{i,t}$ is a hindsight-conditioned compatibility signal rather than an
unbiased or causal estimate of turn-level credit. The analysis does not
guarantee semantic correctness of the generated guidance, monotonic improvement
in expected return, or convergence of the joint task--reflection optimization.

\section{Dataset Details}
\label{app:datasets}

We evaluate embodied reasoning, search-augmented question answering, and web
navigation. Table~\ref{tab:datasets} summarizes the training pools and evaluation sets for the three environments.

\begin{table}[t]
\centering
\renewcommand{\arraystretch}{1.12}
\begin{tabular}{@{}p{0.20\columnwidth}p{0.335\columnwidth}p{0.335\columnwidth}@{}}
\toprule
\raggedright Benchmark
& \raggedright Training pool
& \raggedright Evaluation set\tabularnewline
\midrule
\raggedright ALFWorld
& \raggedright 3,553 tasks from the GiGPO training split
& \raggedright 274 tasks \\ 
(140 seen and 134 unseen)\tabularnewline
\addlinespace
\raggedright Search-based QA
& \raggedright 169,615 questions from NQ and HotpotQA
& \raggedright 51,713 questions across seven datasets \\
(two in-domain and five out-of-domain)\tabularnewline
\addlinespace
\raggedright WebShop
& \raggedright 1,000 tasks
& \raggedright 128 held-out tasks\tabularnewline
\bottomrule
\end{tabular}
\caption{Training-pool sizes and evaluation sets. Training counts denote
entries available for sampling rather than cumulative task instances or
rollout trajectories consumed during optimization.}
\label{tab:datasets}
\end{table}

\paragraph{ALFWorld.}
ALFWorld~\citep{shridhar2020alfworld} converts ALFRED household tasks into a
text interface in which an agent follows observations and admissible commands
to satisfy a natural-language goal. Following SDAR~\citep{lu2026self}, we use
the GiGPO split~\citep{feng2026group}, which contains 3,553 training tasks and
274 evaluation tasks (140 seen and 134 unseen). The six task families are Pick,
Look, Clean, Heat, Cool, and Pick2. We report the success rate for each family
and their weighted average.

\paragraph{Search-based QA.}
Following Search-R1~\citep{jin2025search}, the agent iteratively queries a
retriever and integrates the returned evidence before producing an answer. The
169,615-question training pool contains 79,168 NQ
questions~\citep{kwiatkowski2019natural} and 90,447 HotpotQA
questions~\citep{yang2018hotpotqa}. NQ and HotpotQA also constitute the two
in-domain evaluation sets, while TriviaQA~\citep{joshi2017triviaqa},
PopQA~\citep{mallen2023not}, 2Wiki~\citep{ho2020constructing},
MuSiQue~\citep{trivedi2022musique}, and
Bamboogle~\citep{press2023measuring} provide five out-of-domain evaluation
sets. The seven fixed evaluation sets contain 51,713 questions in total, as
detailed in Table~\ref{tab:searchqa_splits}. We normalize the predicted answer
enclosed by \texttt{<answer>} by lowercasing and removing punctuation,
articles, and extra whitespace. We then report exact-match accuracy for each
dataset and the micro average.

\begin{table}[t]
\centering
\begin{tabular}{llrr}
\toprule
Split & Dataset & Questions & Setting\\
\midrule
Train & NQ          & 79,168 & --\\
      & HotpotQA    & 90,447 & --\\
\cmidrule(lr){2-4}
      & Total       & 169,615 & \\
\midrule
Evaluation & NQ          & 3,610  & ID\\
           & HotpotQA    & 7,405  & ID\\
           & TriviaQA    & 11,313 & OOD\\
           & PopQA       & 14,267 & OOD\\
           & 2Wiki       & 12,576 & OOD\\
           & MuSiQue     & 2,417  & OOD\\
           & Bamboogle   & 125    & OOD\\
\cmidrule(lr){2-4}
           & Total       & 51,713 & \\
\bottomrule
\end{tabular}
\caption{Question counts in the Search-based QA training pool and evaluation
suite. ID and OOD denote in-domain and out-of-domain evaluation, respectively.}
\label{tab:searchqa_splits}
\end{table}

\paragraph{WebShop.}
WebShop~\citep{yao2022webshop} requires an agent to search a product catalog,
inspect product attributes, select appropriate options, and purchase an item
that satisfies a natural-language request. We use 1,000 training tasks and a
fixed held-out evaluation set of 128 tasks. We report the normalized task score
and success rate, both expressed as percentages.

\paragraph{Training and evaluation configuration.}
We train a separate policy per environment. The ALFWorld runs with Qwen3-1.7B
use 600 policy updates, whereas all other runs use 240 updates. With batch
sizes of 16, 16, and 128, the 240-update runs draw 3,840 ALFWorld, 3,840
WebShop, and 30,720 Search-based QA task instances, respectively; the
600-update ALFWorld runs draw 9,600 instances. These are cumulative draws from
the training pools, so repeated tasks are possible. Each sampled task produces
$G=8$ rollouts. Evaluation uses the fixed held-out sets; pooled metrics weight
each evaluated instance equally, while subset columns are reported separately.

\FloatBarrier
\section{Implementation Details}
\label{app:impl}

\begin{table}[t]
\centering
\small
\resizebox{\columnwidth}{!}{%
\begin{tabular}{ll}
\toprule
Hyperparameter & Value\\
\midrule
\multicolumn{2}{l}{\textit{Shared (GRPO algorithm)}}\\
Optimizer                       & AdamW\\
Policy updates                  & $600$ / $240$\\
Learning rate                   & $1\text{e}{-6}$\\
GRPO group size $G$             & $8$\\
PPO clip ratio                  & $0.2$\\
KL loss coefficient             & $0.01$ (low-variance KL)\\
PPO mini-batch size             & $256$ turns\\
ALFWorld      & $2048$ / $512$ tokens\\
SearchQA/WebShop & $4096$ / $512$ tokens\\
Tasks per batch (A/S/W)         & $16$ / $128$ / $16$\\
Max environment steps (A/S/W)  & $50$ / $4$ / $15$\\
Invalid-action penalty (A/S/W)  & $0.1$ / $0.01$ / $0.1$\\
Rollout engine / TP size        & vLLM / $2$\\
Rollout / log-prob micro-batch  & $32$ / $32$ sequences per GPU\\
Rollout maximum model length    & $8608$ tokens \\
Validation temperature          & $0.6$\\
\midrule
\multicolumn{2}{l}{\textit{Baseline-specific (following SDAR)}}\\
OPSD loss coefficient           & $0.01$\\
Skill-SD loss coefficient       & $0.001$\\
RLSD distillation strength      & $0.5$\\
SDAR coefficient / gate $\beta$ & $0.01$ / $5.0$\\
\midrule
\multicolumn{2}{l}{\textit{\textsc{GRSD}-specific (Ours)}}\\
Self-distillation strength $\lambda$   & $0.5$\\
Clip bound $\widehat{G}$               & $0.2$\\
Reflection-loss weight $\alpha$        & $0.01$\\
Reflection / teacher prompt length      & $8096$ / $4096$ tokens\\
Modulation level  & turn\\
Judge temperature / max tokens         & $0.0$ / $16$\\
Judge maximum concurrency               & $16$\\
Judge rubric range                     & $\{0,1,2,3\}$\\
\bottomrule
\end{tabular}
}
\caption{Training hyperparameters. Shared and baseline-specific settings follow
SDAR; \textsc{GRSD} introduces only the configurations in the final block.}
\label{tab:hparams}
\end{table}

\paragraph{Training and Evaluating setup.}
All methods use veRL with Ray-based rollout orchestration, FSDP policy sharding,
and vLLM generation on one node with eight NVIDIA H100 GPUs and tensor-parallel size $2$.
Qwen3-1.7B is trained for 600 policy updates on ALFWorld; all other settings use
240 updates. Within each environment--backbone setting, all methods share the
backbone, environment wrapper, rollout budget, optimizer, and schedule, and
differ only in their method-specific components. We use dynamic batching,
gradient checkpointing, environment seed $0$, and identical task filtering,
action projection, and invalid-action handling across methods.
Table~\ref{tab:hparams} summarizes the hyperparameter settings. Except for the
reported training horizons, shared and baseline-specific configurations follow
SDAR~\citep{lu2026self}; the final block lists the additions specific to
\textsc{GRSD}.

\paragraph{Environment and optimization settings.}
ALFWorld uses 16 tasks per batch, 2,048/512 prompt/response tokens, and at most
50 environment steps. Search-based QA follows Search-R1 with E5 retrieval, 128
tasks per batch, 4,096/512 tokens, and at most four steps. WebShop uses a
1,000-task training pool, 16 tasks per batch, 4,096/512 tokens, and at most 15
steps; its evaluation set contains 128 fixed tasks. Every prompt uses $G=8$
rollouts. We optimize with GRPO, PPO clipping, and a low-variance reference KL
penalty, and sample validation rollouts at temperature $0.6$.

\paragraph{Reflection and guidance generation.}
The policy's vLLM engine generates both the trainable Stage~A reflection $s_i$
and the Stage~B group-reflective guidance $z_x$. Stage~A is optimized by
$\mathcal{L}_{\mathrm{ref}}$, while Stage~B uses the parameter-identical
detached policy and supplies text context only, so no gradient flows through
$z_x$. Trajectories are serialized by turn, with each observation truncated to
400 characters. The policy produces three numbered reflection points, and the
detached snapshot contrasts successful and failed reflections into
two-section DO/AVOID guidance. Single-outcome groups remain inactive and retain
the original GRPO advantage.

\paragraph{Reflection judge.}
A fixed DeepSeek-V4-Flash rubric judge $\mathcal J$ assigns each reflection a
score in $\{0,1,2,3\}$ using the task prompt, verified outcome, trajectory, and
reflection. The rubric assesses causal correctness, outcome consistency,
trajectory faithfulness, and actionability. Scores are group-normalized into
$A_i^{\mathrm{ref}}$; failed or unparseable outputs are masked. Calls use
temperature $0$, at most 16 generated tokens, and concurrency 16. The judge is
used only during training and is absent at inference.

\section{GRSD Algorithm}
\label{app:algorithm}

Algorithm~\ref{alg:grsd} summarizes one \textsc{GRSD} policy update. We first
cache the behavior policy $\pi_{\theta_{\mathrm{old}}}$ and construct its
stop-gradient snapshot $\pi_{\bar\theta}$ for guidance generation and turn-level
re-scoring. For each prompt, the behavior policy samples $G$ trajectories, and
the verifier provides task rewards $R_i$ and binary outcome labels $b_i$. The
rewards are normalized within the rollout group to obtain trajectory-level
advantages $A_i$. In Stage~A, the policy generates one outcome-conditioned
reflection per trajectory, which is scored by the detached judge. Invalid or
unparseable reflections are excluded from reflection-reward normalization and
assigned $A_i^{\mathrm{ref}}=0$.

Stage~B is activated only when the valid reflections include both successful
and failed trajectories. The frozen snapshot contrasts the two reflection sets
to synthesize group-reflective guidance $z_x$. If the required contrast or
valid action turns are unavailable, the original advantages are retained.
Otherwise, the snapshot evaluates each sampled turn under plain and
guidance-augmented contexts. The resulting teacher--student likelihood
difference is normalized and clipped to obtain a bounded turn-level modulation
of $A_i$. Finally, the task, reference-policy KL, and reflection losses are
jointly optimized in one backward pass. Guidance construction and turn
re-scoring remain detached, and only the plain task prompt is used at inference.

\begin{algorithm}[t]
\caption{One GRSD Policy Update}
\label{alg:grsd}
\small
\begin{algorithmic}[1]
\REQUIRE prompt batch $\mathcal B$; policy $\pi_\theta$; reference policy
$\pi_{\mathrm{ref}}$; verifier $R$; judge $\mathcal J$; prompts
$p_{\mathrm{ref}},p_{\mathrm{guide}}$
\REQUIRE $G,\lambda,\widehat G,\alpha,\beta,
\epsilon_{\mathrm{adv}},\epsilon_{\mathrm{gap}}$ with
$\lambda\widehat G<1$
\ENSURE updated policy $\pi_\theta$

\STATE $\theta_{\mathrm{old}}\leftarrow\theta$;
$\bar\theta\leftarrow\operatorname{sg}(\theta_{\mathrm{old}})$

\FOR{each prompt $x\in\mathcal B$}
\STATE Sample $\{\tau_i\}_{i=1}^{G}\sim
\pi_{\theta_{\mathrm{old}}}(\cdot\mid x)$
\STATE $R_i\leftarrow R(\tau_i)$; obtain $b_i\in\{0,1\}$;
$A_i\leftarrow\operatorname{Norm}_{G}(R_i)$ for all $i$

\STATE \textbf{Stage A: policy-native reflection}
\FOR{$i=1,\ldots,G$}
\STATE $s_i\sim\pi_{\theta_{\mathrm{old}}}
(\cdot\mid x,\tau_i,b_i,p_{\mathrm{ref}})$
\STATE $r_i^{\mathrm{ref}}\leftarrow
\operatorname{sg}(\mathcal J(s_i,\tau_i,b_i))$ if the reflection and
judge output are valid; otherwise $r_i^{\mathrm{ref}}\leftarrow\bot$
\ENDFOR
\STATE $\mathcal V_x^{\mathrm{ref}}\leftarrow
\{i:r_i^{\mathrm{ref}}\neq\bot\}$
\STATE $A_i^{\mathrm{ref}}\leftarrow
\operatorname{Norm}_{\mathcal V_x^{\mathrm{ref}}}(r_i^{\mathrm{ref}})$
for $i\in\mathcal V_x^{\mathrm{ref}}$;
$A_i^{\mathrm{ref}}\leftarrow0$ otherwise

\STATE \textbf{Stage B: group contrast and turn re-scoring}
\STATE $\mathcal S_x^+\leftarrow
\{s_i\mid i\in\mathcal V_x^{\mathrm{ref}},\,b_i=1\}$;
$\mathcal S_x^-\leftarrow
\{s_i\mid i\in\mathcal V_x^{\mathrm{ref}},\,b_i=0\}$
\STATE $\mathcal T_x\leftarrow
\{(i,t):1\leq i\leq G,\,1\leq t\leq T_i,\,L_{i,t}>0\}$;
initialize $\widehat A_{i,t}\leftarrow A_i$

\IF{$|\mathcal S_x^+|>0$, $|\mathcal S_x^-|>0$, and
$|\mathcal T_x|>0$}
\STATE $z_x\sim\pi_{\bar\theta}(\cdot\mid
x,\mathcal S_x^+,\mathcal S_x^-,p_{\mathrm{guide}})$
\FOR{each $(i,t)\in\mathcal T_x$}
\STATE $\Delta_{i,t}\leftarrow
\ell^{\mathrm{tea}}_{i,t}(\pi_{\bar\theta},z_x)
-\ell^{\mathrm{stu}}_{i,t}(\pi_{\bar\theta})$
\STATE $\widetilde u_{i,t}\leftarrow
\exp\!\left(\mathrm{sign}(A_i)\Delta_{i,t}\right)-1$
\ENDFOR
\STATE $D_x\leftarrow|\mathcal T_x|^{-1}
\sum_{(j,t')\in\mathcal T_x}|\widetilde u_{j,t'}|
+\epsilon_{\mathrm{gap}}$
\FOR{each $(i,t)\in\mathcal T_x$}
\STATE $\widehat q_{i,t}\leftarrow
\mathrm{clip}(\widetilde u_{i,t}/D_x,-\widehat G,\widehat G)$
\STATE $\widehat A_{i,t}\leftarrow
A_i(1+\lambda\widehat q_{i,t})$
\ENDFOR
\ENDIF
\ENDFOR

\STATE $\mathcal L(\theta)\leftarrow
\mathcal L_{\mathrm{task}}(\theta)
+\beta\mathcal L_{\mathrm{KL}}(\theta;\pi_{\mathrm{ref}})
+\alpha\mathcal L_{\mathrm{ref}}(\theta)$
\STATE $\theta\leftarrow\operatorname{OptStep}
(\theta,\mathcal L(\theta))$
\end{algorithmic}
\end{algorithm}

\section{Computational Cost}
\label{app:cost}

\begin{table*}[t]
\centering
\small
\setlength{\tabcolsep}{10pt}
\begin{tabular}{lccc}
\toprule
Component & GRPO & SDAR & \textsc{GRSD}\\
\midrule
\multicolumn{4}{l}{\textbf{\textit{Additional operations}}}\\
Guidance source
  & -- & cached offline skills & online group reflections\\
Teacher forward passes
  & -- & $1$ per prompt & $1$ per prompt\\
Reflection generation
  & -- & -- & $G$ per prompt\\
Group-level synthesis
  & -- & -- & $1$ per prompt\\
Reflection-loss optimization
  & -- & -- & joint with task update\\
External rubric calls
  & -- & -- & $G$ per prompt\\
\midrule
\multicolumn{4}{l}{\textbf{\textit{Average wall-clock time per update (s)}}}\\
Rollout generation
  & 252.8 & 251.3 & 251.6\\
Teacher forward
  & -- & 25.1 & 12.0\\
Guidance construction
  & -- & -- & 29.4\\
Reflection-loss computation
  & -- & -- & 7.3\\
Other operations
  & 31.3 & 31.5 & 33.1\\
\midrule
Total training
  & 284.1 & 307.9 & 333.4\\
Relative cost
  & 1.00$\times$ & 1.08$\times$ & 1.17$\times$\\
\bottomrule
\end{tabular}
\caption{Additional operations and observed training cost on ALFWorld with
Qwen3-1.7B. Wall-clock times are reported in seconds per update and averaged over all 600 policy updates, excluding validation and checkpoint I/O.}
\label{tab:cost}
\end{table*}

Compared with GRPO, SDAR retrieves cached offline skills and performs one
skill-conditioned teacher forward, without auxiliary generation or a
separate optimization step. In contrast, \textsc{GRSD} constructs guidance
online. For each prompt, Stage~A generates one reflection for each of the $G$
rollouts, Stage~B synthesizes these reflections once at the group level, and
the resulting guidance conditions a forward-only teacher pass that re-scores
the sampled turns. Optimizing the Stage~A reflections introduces additional
reflection-loss computation, which is jointly performed with task optimization
in the same backward pass and policy update. Both reflection and guidance
generation reuse the policy's vLLM engine, while the external rubric judge is
invoked only during training. Therefore, none of these additional operations
incurs inference-time overhead, and all three methods use the same plain-prompt
inference procedure.

As shown in Table~\ref{tab:cost}, \textsc{GRSD} increases the observed
per-update training time by $17\%$ over GRPO, compared with an $8\%$ increase
for SDAR. Most of the additional cost arises from online guidance construction,
whereas reflection-loss computation contributes a relatively small fraction.
Notably, although SDAR and \textsc{GRSD} each require one teacher forward pass,
the latter takes only $12.0$ seconds per update, compared with $25.1$ seconds
for SDAR. This reduction is consistent with Stage~B distilling rollout-level
reflections into compact, prompt-specific guidance, thereby shortening the
teacher pass and partially offsetting the cost of online guidance construction
relative to conditioning on retrieved offline skills. Overall, the modest
training-only overhead of \textsc{GRSD}, together with its larger performance
improvements reported in the main results, indicates a favorable
performance--cost trade-off.

\section{Group-Reflective Skill Examples}
\label{app:skill_examples}

To make the construction concrete, we present a recorded ALFWorld group for the
task ``cool some mug and put it in coffeemachine''. The group contains three
successful and five failed rollouts. The example covers every rollout in the
group: each Stage~A reflection is abridged for readability, while the complete
Stage~B guidance is reproduced verbatim apart from typographic formatting.
Parentheses report the reflection-judge score $\mathcal J$. Stage~B constructs
the guidance from the unabridged reflections of all eight rollouts; the judge
scores provide learning signals for Stage~A and are not inputs to guidance
construction.

{\small
\noindent\textbf{Stage A: successful trajectory reflections (abridged).}
\begin{enumerate}[label=\textbf{S\arabic*.},leftmargin=2.1em,
                  topsep=3pt,itemsep=2pt,parsep=0pt,partopsep=0pt]
  \item $\mathcal J=1$. Located and took a mug on countertop~1; heated it in
        the microwave ``to prepare it for cooling''; cooled it in the fridge;
        and moved it to coffeemachine~1.
  \item $\mathcal J=2$. Examined countertop~1 and reported the mug absent; then
        cooled the mug with fridge~1 and moved it to coffeemachine~1.
  \item $\mathcal J=2$. Located mug~3 on countertop~2, cooled it with fridge~1,
        and moved it directly to coffeemachine~1 while avoiding redundant
        fridge actions.
\end{enumerate}

\noindent\textbf{Stage A: failed trajectory reflections (abridged).}
\begin{enumerate}[label=\textbf{F\arabic*.},leftmargin=2.1em,
                  topsep=3pt,itemsep=2pt,parsep=0pt,partopsep=0pt]
  \item $\mathcal J=3$. Repeatedly searched the countertop, bowl, and fridge
        without locating the mug, and unnecessarily attempted to cool the bowl.
  \item $\mathcal J=2$. Cooled apple~1 and placed it in the coffeemachine, but
        never located or cooled the target mug.
  \item $\mathcal J=2$. Repeatedly cooled a bowl and moved it to the
        coffeemachine, then incorrectly claimed that this wrong-object sequence
        completed the mug task.
  \item $\mathcal J=0$. Mixed claims about locating and cooling a mug with
        actions on apple~2, and incorrectly claimed that placing the apple
        completed the task.
  \item $\mathcal J=1$. Repeatedly searched the countertop and bowl, failed to
        locate the mug, and recommended searching a cabinet or fridge more
        directly.
\end{enumerate}

\noindent\textbf{Stage B: synthesized group-reflective guidance $z_x$ (complete).}

\smallskip
\noindent\textbf{Key points to DO}
\begin{itemize}[leftmargin=1.25em,topsep=2pt,itemsep=1pt,parsep=0pt,partopsep=0pt]
  \item \textbf{Locate the mug} on countertop~1 or~2 before attempting to cool it.
  \item \textbf{Use the fridge} to cool the mug, as it is the most direct and
        reliable method for cooling.
  \item \textbf{Move the cooled mug} to coffeemachine~1 directly after cooling,
        without unnecessary steps.
\end{itemize}

\noindent\textbf{Mistakes to AVOID}
\begin{itemize}[leftmargin=1.25em,topsep=2pt,itemsep=1pt,parsep=0pt,partopsep=0pt]
  \item \textbf{Fail to locate the mug} in the environment, leading to repeated
        unsuccessful attempts.
  \item \textbf{Attempt to cool the bowl} instead of the mug, which is irrelevant
        to the task.
  \item \textbf{Retry actions without a clear goal}, such as repeatedly opening
        and closing the fridge, which wastes time and resources.
\end{itemize}
}

This complete group makes the contrast explicit: successful reflections share
the locate--cool--place sequence, whereas failures repeatedly lose the target or
act on a bowl or apple.  Despite noise in individual reflections, the synthesized
guidance retains the common successful sequence and recurring failure modes.

\section{Additional Results}
\label{app:additional_results}

\begin{figure*}[t]
\centering

\begin{minipage}[t]{0.32\textwidth}
    \centering
    \includegraphics[width=\linewidth]{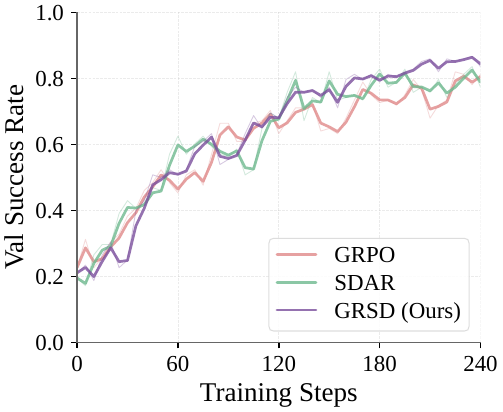}
    \vspace{-0.6em}

    {\small (a) ALFWorld\par}
\end{minipage}
\hfill
\begin{minipage}[t]{0.32\textwidth}
    \centering
    \includegraphics[width=\linewidth]{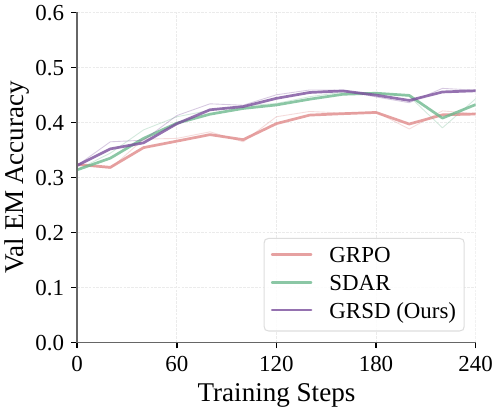}
    \vspace{-0.6em}

    {\small (b) Search-based QA\par}
\end{minipage}
\hfill
\begin{minipage}[t]{0.32\textwidth}
    \centering
    \includegraphics[width=\linewidth]{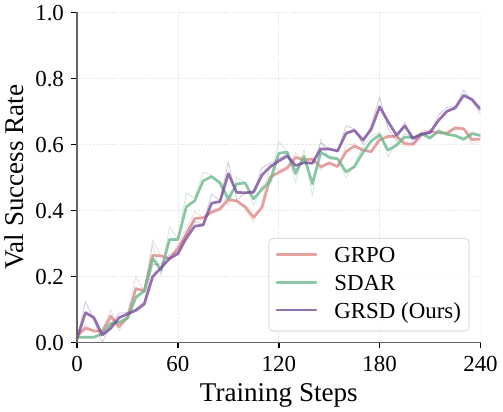}
    \vspace{-0.6em}

    {\small (c) WebShop\par}
\end{minipage}

\vspace{-0.5em}
\caption{Validation success over 240 training steps with
Qwen2.5-3B-Instruct across three environments.}
\label{fig:qwen25_3b_training_dynamics}
\end{figure*}

\begin{figure*}[t]
\centering

\begin{minipage}[t]{0.32\textwidth}
    \centering
    \includegraphics[width=\linewidth]{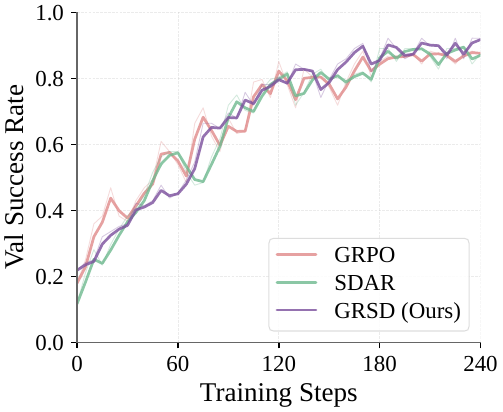}
    \vspace{-0.6em}

    {\small (a) ALFWorld\par}
\end{minipage}
\hfill
\begin{minipage}[t]{0.32\textwidth}
    \centering
    \includegraphics[width=\linewidth]{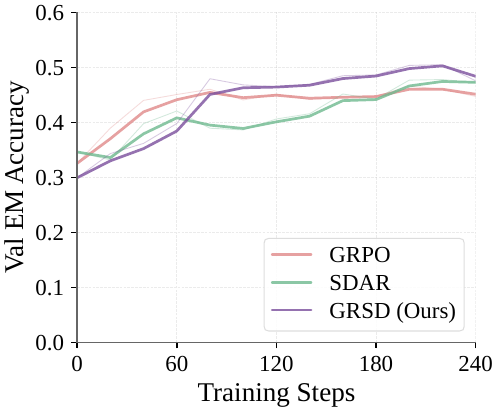}
    \vspace{-0.6em}

    {\small (b) Search-based QA\par}
\end{minipage}
\hfill
\begin{minipage}[t]{0.32\textwidth}
    \centering
    \includegraphics[width=\linewidth]{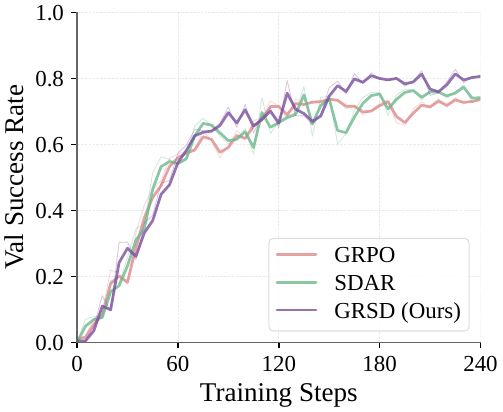}
    \vspace{-0.6em}

    {\small (c) WebShop\par}
\end{minipage}

\vspace{-0.5em}
\caption{Validation success over 240 training steps with
Qwen2.5-7B-Instruct across three environments.}
\label{fig:qwen25_7b_training_dynamics}
\end{figure*}

We provide additional analyses of GRSD's training behavior and scalability.
We first examine the learning dynamics of policy-native reflections in
Stage~A and the availability of mixed-outcome rollout groups for contrastive
guidance construction in Stage~B. We then compare validation performance
across environments and model scales to assess whether the benefits of
group-reflective guidance persist with stronger backbones. Translucent curves denote raw measurements and solid curves denote smoothed trends.

\begin{figure}[t]
\centering

\begin{minipage}[t]{0.49\linewidth}
    \centering
    \includegraphics[width=\linewidth]{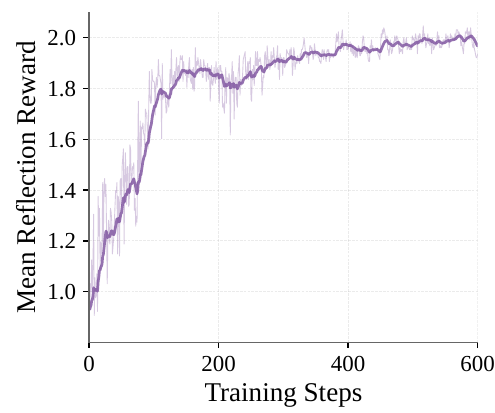}
    \vspace{-0.6em}

    {\small (a) Mean reflection reward\par}
\end{minipage}
\hfill
\begin{minipage}[t]{0.49\linewidth}
    \centering
    \includegraphics[width=\linewidth]{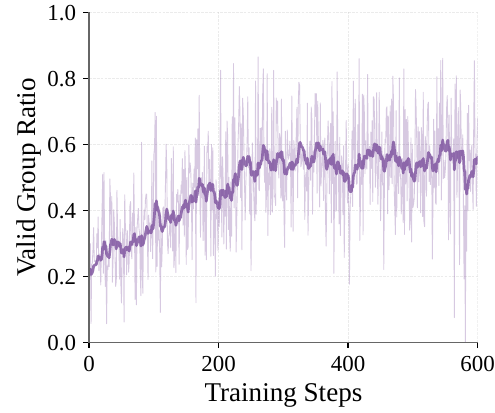}
    \vspace{-0.6em}

    {\small (b) Valid-group ratio\par}
\end{minipage}

\vspace{-0.5em}
\caption{Training dynamics on ALFWorld. (a) Mean judge-assigned
reward of the policy-native reflections generated in Stage~A. (b) Proportion
of rollout groups containing both successful and failed trajectories and thus
eligible for contrastive guidance construction in Stage~B.}
\label{fig:reflection_group_dynamics}
\end{figure}

\subsection{Policy-Native Reflection Learning}
\label{app:reflection_learning_dynamics}

Unlike the task reward, the reflection reward in
Figure~\ref{fig:reflection_group_dynamics}(a) evaluates the outcome consistency,
causal grounding, and actionability of model-generated reflections. The mean
reward rises rapidly early in training, suggesting that the policy quickly
learns to identify outcome-critical behaviors and provide actionable feedback.
After brief fluctuations, it converges to a substantially higher level,
indicating sustained improvement in the quality of policy-native reflections.

\subsection{Valid-Group Dynamics}
\label{app:valid_group_dynamics}

A rollout group is valid for Stage~B if it contains both successful and failed
trajectories. As shown in Figure~\ref{fig:reflection_group_dynamics}(b), the
valid-group ratio rises from a low initial level and then remains within a
nontrivial intermediate range despite per-step fluctuations. This pattern shows
that mixed-outcome groups remain available as the policy improves, sustaining
the success--failure contrasts required for guidance construction. Together
with the increasing reflection reward, these results indicate that Stage~B
receives both persistent contrastive evidence and increasingly informative
trajectory reflections. For invalid groups, GRSD retains the original GRPO
advantage without guidance-based modulation.

\subsection{Training Dynamics Across Model Scales}
\label{app:backbone_training_dynamics}

Figures~\ref{fig:qwen25_3b_training_dynamics} and
\ref{fig:qwen25_7b_training_dynamics} compare validation dynamics across
ALFWorld, Search-based QA, and WebShop. At both model scales, \textsc{GRSD}
progressively establishes a clear performance advantage and achieves the best
final results across all three environments, with particularly pronounced gains
on ALFWorld and WebShop. Its consistent superiority on both the 3B and 7B
backbones demonstrates that group-reflective guidance remains effective as model
capacity increases.

\section{Prompt Templates}
\label{app:prompts}

We provide the full prompt templates used in our experiments. Braces denote
runtime-filled slots. During training, the privileged self-teacher additionally
prepends the group-reflective skill guidance $z_x$ to the agent context; at
inference the agent sees only the plain prompts below.
For the teacher forward pass, the per-group guidance is prepended to the original
prompt as the following text block. This prefix is used only to compute teacher
log probabilities and is detached from the policy update; it never changes the
sampled response or trajectory and is not supplied at inference.

\subsection{Agent Prompts}

The environment-facing agent prompts are given in paired initial-observation
and subsequent-observation forms. The ALFWorld pair appears first, followed by
the Search-Based QA and WebShop pairs. Each pair shares the same action-format
requirements while filling its environment-specific observation and history
slots.

\subsection{Self-Reflection Prompt ($p_{\mathrm{ref}}$)}

The policy completes the corresponding panel below to produce the trainable
reflection $s_i$. The outcome and trajectory slots are filled from one
verified rollout, and the output is restricted to three grounded imperative
points.

\subsection{Guidance Construction Prompt ($p_{\mathrm{guide}}$)}

The parameter-identical detached policy branch synthesizes the group's
reflections into the contrastive DO/AVOID guidance $z_x$. The corresponding panel
receives separate successful and failed reflection blocks and is used only for
mixed-outcome groups.

\subsection{Reflection Judge Prompt ($\mathcal{J}$)}

The external judge scores the reflection panel's output on a $\{0,1,2,3\}$
rubric. Hidden reasoning is disabled and the judge returns only one digit.

% Full-width, page-breakable prompt panels used by Appendix~\ref{app:prompts}.
%
% Required in the main-document preamble:
%   \usepackage{xcolor}
%   \usepackage{listings}
%   \usepackage[most]{tcolorbox}
%
% The former cuted/strip implementation is intentionally removed. Each
% promptpanel is now a breakable tcolorbox in one-column mode.

\definecolor{promptpanelaccent}{RGB}{45,91,96}
\definecolor{promptpanelborder}{RGB}{185,193,197}
\definecolor{promptpaneltitle}{RGB}{239,243,243}
\definecolor{promptpaneltext}{RGB}{33,37,41}

\lstdefinestyle{prompttemplate}{
  basicstyle=\rmfamily\normalsize\color{promptpaneltext},
  columns=fullflexible,
  keepspaces=true,
  upquote=true,
  showstringspaces=false,
  breaklines=true,
  breakatwhitespace=true,
  breakindent=1.5em,
  frame=none,
  aboveskip=0pt,
  belowskip=0pt,
  xleftmargin=0pt,
  xrightmargin=0pt
}

\newtcblisting{promptpanel}[1]{%
  enhanced jigsaw,
  breakable,
  listing only,
  listing engine=listings,
  listing options={style=prompttemplate},
  title={#1},
  title after break={#1\space\textit{(continued)}},
  fonttitle=\bfseries\normalsize,
  coltitle=promptpaneltext,
  colbacktitle=promptpaneltitle,
  colback=white,
  colframe=promptpanelborder,
  boxrule=0.5pt,
  titlerule=0.5pt,
  borderline west={2.2pt}{0pt}{promptpanelaccent},
  arc=1mm,
  outer arc=1mm,
  left=11pt,
  right=11pt,
  top=9pt,
  bottom=9pt,
  toptitle=7pt,
  bottomtitle=7pt,
  before skip=8pt,
  after skip=8pt,
  pad at break*=1mm
}

% Switch to a normal one-column page so the full-width boxes can break
% naturally across page boundaries.
\onecolumn

% Uncomment this panel if the privileged teacher prefix should be displayed.
%
\begin{promptpanel}{Privileged Teacher Context}
[Privileged Skill Information]
{group_reflective_guidance}

{original_task_prompt}
\end{promptpanel}

\begin{promptpanel}{ALFWorld Prompt: Initial Observation}
You are an expert agent operating in the ALFRED Embodied Environment.
Your current observation is: {current_observation}
Your admissible actions of the current situation are: [{admissible_actions}].

Now it's your turn to take an action.
You should first reason step-by-step about the current situation. This reasoning process MUST be enclosed within <think> </think> tags.
Once you've finished your reasoning, you should choose an admissible action for current step and present it within <action> </action> tags.
\end{promptpanel}

\begin{promptpanel}{ALFWorld Prompt: Subsequent Observation}
You are an expert agent operating in the ALFRED Embodied Environment. Your task is to: {task_description}
Prior to this step, you have already taken {step_count} step(s). Below are the most recent {history_length} observations and the corresponding actions you took:
{action_history}
You are now at step {current_step} and your current observation is: {current_observation}
Your admissible actions of the current situation are: [{admissible_actions}].

Now it's your turn to take an action.
You should first reason step-by-step about the current situation. This reasoning process MUST be enclosed within <think> </think> tags.
Once you've finished your reasoning, you should choose an admissible action for current step and present it within <action> </action> tags.
\end{promptpanel}

\begin{promptpanel}{Search-Based QA Prompt: Initial Observation}
You are an expert agent tasked with answering the given question step-by-step.
Your question: {task_description}

Now it's your turn to respond for the current step.
You should first conduct reasoning process. This process MUST be enclosed within <think> </think> tags.
After completing your reasoning, choose only one of the following actions (do not perform both):
(1) If you find you lack some knowledge, you can call a search engine to get more external information using format: <search> your query </search>.
(2) If you have enough knowledge to answer the question confidently, provide your final answer within <answer> </answer> tags, without detailed illustrations. For example, <answer>Beijing</answer>.
\end{promptpanel}

\begin{promptpanel}{Search-Based QA Prompt: Subsequent Observation}
You are an expert agent tasked with answering the given question step-by-step.
Your question: {task_description}

Prior to this step, you have already taken {step_count} step(s). Below is the interaction history where <search> </search> wrapped your past search queries and <information> </information> wrapped the search results returned by the engine.
History: {memory_context}

Now it's your turn to respond for the current step.
You should first conduct reasoning process. This process MUST be enclosed within <think> </think> tags.
After completing your reasoning, choose only one of the following actions (do not perform both):
(1) If you find you lack some knowledge, you can call a search engine to get more external information using format: <search> your query </search>.
(2) If you have enough knowledge to answer the question confidently, provide your final answer within <answer> </answer> tags, without detailed illustrations. For example, <answer>Beijing</answer>.
\end{promptpanel}

\begin{promptpanel}{WebShop Prompt: Initial Observation}
You are an expert autonomous agent operating in the WebShop e-commerce environment.
Your task is to: {task_description}.
Your current observation is: {current_observation}.
Your admissible actions of the current situation are: [{available_actions}].

Now it's your turn to take one action for the current step. You should first reason step-by-step about the current situation, then think carefully which admissible action best advances the shopping goal. This reasoning process MUST be enclosed within <think> </think> tags.
Once you've finished your reasoning, you should choose an admissible action for current step and present it within <action> </action> tags.
\end{promptpanel}

\begin{promptpanel}{WebShop Prompt: Subsequent Observation}
You are an expert autonomous agent operating in the WebShop e-commerce environment.
Your task is to: {task_description}.
Prior to this step, you have already taken {step_count} step(s). Below are the most recent {history_length} observations and the corresponding actions you took: {action_history}
You are now at step {current_step} and your current observation is: {current_observation}.
Your admissible actions of the current situation are: [{available_actions}].

Now it's your turn to take one action for the current step. You should first reason step-by-step about the current situation, then think carefully which admissible action best advances the shopping goal. This reasoning process MUST be enclosed within <think> </think> tags.
Once you've finished your reasoning, you should choose an admissible action for current step and present it within <action> </action> tags.
\end{promptpanel}

\begin{promptpanel}{Stage A: Policy-Native Self-Reflection}
[System]
You are the same agent that just attempted an interactive task. Reflect on your own trajectory and distill a short, concrete strategy you can reuse. Be specific and grounded in the decisions, actions, and evidence observed.

[User]
Reflect on ONE of your own completed trajectories.

# Task
{task_text}

# Outcome
This trajectory {SUCCEEDED | FAILED}.

# Your interaction history (turn by turn)
{trajectory_text}

# Instructions
Write 3 numbered key points that capture the decisive behavior of THIS trajectory.
- If it SUCCEEDED: state concretely what you did RIGHT (the decisions/steps that caused success).
- If it FAILED: state concretely what went WRONG and which action(s) to AVOID or do differently.
Each key point: one imperative sentence grounded in the actual actions. Output only the list.
\end{promptpanel}

\begin{promptpanel}{Stage B: Group-Level Guidance Construction}
[System]
You summarize reflections from several trajectories of the SAME task (some succeeded, some failed) into one compact contrastive guideline.

[User]
Below are reflections from multiple trajectories attempting the SAME task.

# Task
{task_text}

# Reflections from SUCCESSFUL trajectories
{positive_block}

# Reflections from FAILED trajectories
{negative_block}

# Instructions
Induce a compact guideline with EXACTLY these two sections and nothing else:

### Key points to DO (from successful trajectories)
- 3-5 imperative bullet points describing the correct decisions/steps that lead to success.

### Mistakes to AVOID (from failed trajectories)
- 3-5 imperative bullet points describing the wrong actions to avoid, derived from the failures.

Be concrete and task-specific. Output only the two sections with their bullet points.
\end{promptpanel}

\begin{promptpanel}{Reflection Judge}
[System]
You are a strict, discriminating rubric grader for self-reflections written by an interactive task agent. You are given the task, verified outcome, full trajectory, and the agent's reflection. Grade ONLY the quality of the reflection as a causally correct, outcome-consistent, and reusable skill, NOT whether the trajectory itself succeeded. Reserve the top score for flawless reflections; when torn between scores, choose the lower one.

[User]
# Grade the agent's self-reflection using the rubric below.

# Task
{task_text}

# Verified outcome
This trajectory {SUCCEEDED | FAILED}.

# Trajectory (turn by turn)
{trajectory_text}

# Agent's reflection to grade
{reflection_text}

# What to judge (assess all four dimensions)
1. Causal correctness: identify the decisive turns or actions that caused the outcome.
2. Outcome consistency: for success, state what to repeat; for failure, state what went wrong and what to avoid or fix.
3. Faithfulness: every referenced action or step must appear in the trajectory.
4. Actionability: each point must be concrete and reusable rather than a generic platitude.

# Rubric (choose exactly one integer from 0 to 3)
3 = Excellent: decisive, causally correct, grounded, outcome-consistent, and fully actionable.
2 = Good: captures the core decisive behavior but has a minor omission, vague point, or imprecision.
1 = Weak: misses the decisive turn, is mostly generic, or includes unsupported content.
0 = Poor: incorrect, outcome-inconsistent, hallucinated, non-actionable, or off-topic.

# Calibration
- A reasonable but generic reflection is at most 1.
- Do not give 3 if any flaw remains.
- When torn between scores, choose the lower one.

# Output format
Respond with ONLY a single character: 0, 1, 2, or 3.
No words, no punctuation.
\end{promptpanel}

% If more two-column appendix content follows, restore it with:
% \twocolumn

\end{document}